\documentclass[10pt,journal,final]{IEEEtran}

\ifCLASSOPTIONcompsoc
\usepackage[nocompress]{cite}
\else
\usepackage{cite}
\fi

\usepackage{bbding}

\usepackage{amsfonts}

\usepackage{amsmath}

\usepackage{graphicx}

\usepackage{float}  

\usepackage{subfigure}

\usepackage{bm}

\usepackage{courier}

\usepackage{booktabs}

\usepackage{multirow}

\newcommand{\tabincell}[2]{\begin{tabular}{@{}#1@{}}#2\end{tabular}}

\usepackage{rotating}

\usepackage{amsthm}

\usepackage{footnote}

\usepackage{color}

\usepackage{stfloats}

\usepackage{threeparttable}

\usepackage{adjustbox}

\begin{document}

\title{Shrinking the Semantic Gap:\\ Spatial Pooling of Local Moment Invariants for Copy-Move Forgery Detection}

\author{Chao Wang, Zhiqiu Huang, Shuren Qi, Yaoshen Yu, Guohua Shen, and Yushu Zhang,~\IEEEmembership{Member,~IEEE}
\IEEEcompsocitemizethanks{	
	\IEEEcompsocthanksitem This work was supported in part by the Joint Funds of the National Natural Science Foundation of China under Grant U2241216, in part by the Opening Fund of Key Laboratory of Civil Aviation Emergency Science \& Technology under Grant NJ2022022, in part by the State Key Laboratory of Information Security under Grant 2022-MS-02, and in part by the Postgraduate Research \& Practice Innovation Program of Jiangsu Province under Grant KYCX22-0383. \emph{(Corresponding author: Z. Huang}.)
	\IEEEcompsocthanksitem C. Wang, Z. Huang, S. Qi, Y. Yu, and G. Shen are with the College of Computer Science and Technology, Nanjing University of Aeronautics and Astronautics, Nanjing, China (e-mail: c.wang@nuaa.edu.cn, zqhuang@nuaa.edu.cn, shurenqi@nuaa.edu.cn, yaoshen.yu@outlook.com, and ghshen@nuaa.edu.cn).
	\IEEEcompsocthanksitem Y. Zhang is with the College of Computer Science and Technology, Nanjing University of Aeronautics and Astronautics, Nanjing, China, and also with the State Key Laboratory of Information Security, Institute of Information Engineering, Chinese Academy of Sciences, Beijing, China (e-mail: yushu@nuaa.edu.cn).
	
	\IEEEcompsocthanksitem IEEE Transactions on Information Forensics and Security (TIFS), 2023, doi: 10.1109/TIFS.2023.3234861, link: ieeexplore.ieee.org/document/10007894.

	}}

\markboth{Wang \MakeLowercase{\textit{et al.}}: Shrinking the Semantic Gap: Spatial Pooling of Local Moment Invariants for Copy-Move Forgery Detection}
{Wang \MakeLowercase{\textit{et al.}}: Shrinking the Semantic Gap: Spatial Pooling of Local Moment Invariants for Copy-Move Forgery Detection}


\maketitle

\begin{abstract}
Copy-move forgery is a manipulation of copying and pasting specific patches from and to an image, with potentially illegal or unethical uses. Recent advances in the forensic methods for copy-move forgery have shown increasing success in detection accuracy and robustness. However, for images with high self-similarity or strong signal corruption, the existing algorithms often exhibit inefficient processes and unreliable results. This is mainly due to the inherent semantic gap between low-level visual representation and high-level semantic concept. In this paper, we present a very first study of trying to mitigate the semantic gap problem in copy-move forgery detection, with spatial pooling of local moment invariants for midlevel image representation. Our detection method expands the traditional works on two aspects: 1) we introduce the bag-of-visual-words model into this field for the first time, may meaning a new perspective of forensic study; 2) we propose a word-to-phrase feature description and matching pipeline, covering the spatial structure and visual saliency information of digital images. Extensive experimental results show the superior performance of our framework over state-of-the-art algorithms in overcoming the related problems caused by the semantic gap.
\end{abstract}


\section{Introduction}
\IEEEPARstart{I}{n} the last few years, multimedia forensics appears in full development with high demands from industry and society. This is mainly because fake multimedia has become an urgent problem in the real world, and they are often used to manipulate public opinion, commit fraud, discredit or blackmail people.

As a common and basic image manipulation methods,  \emph{copy-move forgery} \cite{ref1} consists of copying and pasting specific patches from and to an image, in order to over-emphasize or cover the objects of interest. Although copy-move forgery can be easily performed by any individual with modern image editing software (e.g., Photoshop), detecting such manipulations is not a trivial task. The fundamental difficulty lies in finding the correspondences among copy-move regions, which exhibit near-identical semantic properties but differ at the digital level due to the trace removal operations such as geometric transformations and signal corruptions.

In general, classical copy-move forgery detection algorithm covers three phases: 1) feature extraction, 2) matching, and 3) post-processing \cite{ref2}. The main ideas are 1) extracting suitable local features to represent the properties of the image patches, 2) finding the best counterpart of each patch by matching the corresponding features, and 3) converting such matches to the final detection result by post-processing.

As a critical phase, the feature extraction for copy-move detection forgery have been pursued by many researchers in two different branches: dense-field and sparse-field \cite{ref3}. Dense-field approach extracts features from overlapping and fixed-size image blocks. For achieving higher robustness, some well-known transformations like moment invariants have been used to construct the block features. Despite their success in ideal scenarios, this line of approaches intrinsically lacks scaling invariance and efficient calculation \cite{ref4}. On the contrary, sparse-field approach shows superiority in solving the above limitations, since the feature extraction works only on a relatively small set of patches (i.e., keypoints) with certain geometric invariance. In this branch, the main problem is lower accurate than the dense-field methods due to the sparse nature, especially when the forgery only involves small or smooth regions \cite{ref4}. It is an obvious fact that above two branches have difficulties in achieving satisfactory detection accuracy, efficiency and robustness at the same time.

Through various differences, we observe the following common phenomena in both state-of-the-art dense-field and sparse-field detection methods: 1) for the image with high self-similarity, they tend to produce large number of false matches due to the poor discriminability of features; 2) for the image with strong signal corruption, they struggle to obtain a sufficient number of correct matches due to the poor robustness of features. Note that such limitations in feature extraction will in turn lead to inefficient matching/post-processing processes and unreliable detection results.

The fundamental cause for such phenomena lies in the well-known \emph{semantic gap}. Specifically, copy-move forgery detection mainly focuses on checking the \emph{semantic integrity}, so it is critical to effectively model local semantic properties from digital images. However, most copy-move forgery detection methods rely heavily on the low-level visual representation of digital images, e.g., Scale Invariant Feature Transform (SIFT) \cite{ref5}, and there is a big gap between low-level features and high-level semantic concepts. With this in mind, one main explanation for the semantic gap is that detection algorithms ignore the \emph{spatial structure relationships} and \emph{visual saliency differences} of the extracted features. In other words, such \emph{independent} and \emph{indifferent} features are not sufficient to model the complex semantic properties.

Motivated by the above facts, this paper attempts to take a first step towards alleviating the semantic gap problem in copy-move forgery detection. Our work has the following two key contributions.
\begin{itemize}
	\item We introduce the \emph{Bag-of-Visual-Words} (BoVW) model \cite{ref47} into the field of copy-move forgery detection, meaning a new theoretical basis for feature extraction. In overcoming the semantic gap, the BoVW model may be the most competitive representation model in pre-deep-learning era \cite{ref6}. It is a hand-crafted representation in which low-level local features are extracted, encoded, and summarized into high-level image features.
	\item We propose a \emph{word-to-phrase feature description and matching pipeline} inspired by BoVW model, covering the spatial structure and visual saliency information of digital images. In this hierarchical strategy, local features are first roughly described and matched at the visual-word level, followed by a fine description and matching of these matched features at the visual-phrase level. The core lies in a spatial pooling (spatial structure) and weighting (visual saliency) of low-level features for the mid-level representation, which is expected to alleviate the semantic gap. 
\end{itemize}

\section{Related Work}
In this section, we will discuss the existing branches for copy-move forgery detection, with special attention on successful experiences and common limitations behind such works.

\subsection{Dense-Field Approach}
In this approach, most of the research efforts have focused improving the robustness of the features and the efficiency of the matching.

For feature extraction, it is a natural requirement that features can be robust against conventional signal corruption and geometric distortion. Some early works adopt classical image transform with high robustness to JPEG compression, noise addition, and other common distortions. The popular choices in the literature include Discrete Cosine Transform (DCT) \cite{ref7}, Wavelet Transform (WT) \cite{ref8}, Fourier Transform (FT) \cite{ref9}, and Singular Value Decomposition (SVD) \cite{ref10}. However, the features extracted by such classical methods are vulnerable to geometric transforms, mainly rotation and scaling. This observation prompts the introduction of more complex geometric invariants. In this regard, a common strategy is to provide the rotation invariance by using the orthogonal moments, e.g., Zernike Moments (ZM) \cite{ref11}, Fourier-Mellin Transform (FMT) \cite{ref12}, Polar Cosine Transform (PCT) \cite{ref13}, and Polar Complex Exponential Transform (PCET) \cite{ref14}. As for the scale invariance, however, effective generation methods are still lacking in dense-field approach. In fact, many scale-invariant transforms (e.g., FMT) require the interest region for feature extraction to be \emph{adaptive} or \emph{infinite}. But the size of the image block is generally fixed in dense-field detection algorithms.

For feature matching, the processing of a large number of blocks (about  $10^5 \sim 10^6$ blocks per image) usually results in expensive calculations, especially the brute-force method. A variety fast implementations for the generic nearest-neighbor (NN) searching have been introduced into this field, such as lexicographic sorting \cite{ref15}, kd-trees \cite{ref16}, and Locality Sensitive Hashing (LSH) \cite{ref11}. However, these methods either cause a significant reduction in robustness/accuracy, or still fail to bring a satisfactory speed. One convincing reason is that such generic algorithms do not fully consider the specific background knowledge (i.e., priori) of the copy-move forgery detection problem. Inspired by this, more efficient NN searching have been proposed, e.g., modified PM \cite{ref17} and enhanced Coherency Sensitive Hashing (CSH) \cite{ref13}, by exploiting the smoothness and self-similarity of the image. Note that, despite their success, such speed-up algorithms still cannot be used in time-critical scenarios, especially when the high-resolution image or high-reliability detection is required.

In summary, high computational burden and poor scaling invariance are the two intractable problems of dense-field detection algorithms. This also promotes the development of sparse-field approach. 

\subsection{Sparse-Field Approach}
Sparse-field approach is an alternative path for copy-move forgery detection with inherent advantages in terms of complexity and invariance. Therefore, in this approach, the design of feature extraction, matching, and post-processing have all attracted extensive research interest.

For feature extraction, regions of interest are first detected (i.e., keypoint detection) and then described by specific representation (i.e., keypoint description). A hot issue in the detection is the lack of keypoints in smooth or small regions, which may also contain copy-move forgery. Existing solutions include threshold adjustment \cite{ref18}, resolution enhancement \cite{ref19}, and Non-Maximum Suppression (NMS) \cite{ref20}. With such strategies, the state-of-the-art forensic methods are able to detect the smooth or small copy-move region more effectively. As for the description, similar to the case of dense-field approach, achieving satisfactory descriptor is a long-lasting battle. The most commonly used SIFT do not perform well in terms of robustness and efficiency \cite{ref21}, especially the invariance to geometric changes. Therefore, the variations of SIFT with more invariance or lower complexity, such as SURF \cite{ref22}, KAZE \cite{ref23}, OpponentSIFT \cite{ref20}, affine-SIFT \cite{ref24}, mirror-SIFT \cite{ref25}, and binarized SIFT \cite{ref26}, were successively introduced into recent works. Also, moment invariants are used to describe the keypoints considering their geometric invariance and compactness, such as Exponent-Fourier Moments (EFM) \cite{ref18}, PCT \cite{ref27}, PCET \cite{ref28}, Radial Harmonic Fourier Moments (RHFM) \cite{ref29}, and Bessel-Fourier Moments (BFM) \cite{ref30}. As the other side of the coin, such moment invariants are less informative than SIFT-like texture features, implying lower discriminative power. Thus, for all the above features, it has difficulties in achieving satisfactory robustness and discriminability at the same time. This phenomenon is in fact a sign of the semantic gap between low-level visual representation and high-level semantic concept.

For feature matching, candidate features are first searched out from the set (i.e., NN searching), and then the true matches are screened out through certain testing (i.e., NN testing). In general, the NN searching dominates the computational cost of the matching stage. Efficiency gains can be achieved by introducing kd-trees \cite{ref31} or hashing \cite{ref26}, similar to the case of dense-field approach. Other researchers have further accelerated the above NN searching methods by features grouping \cite{ref19}, based on certain prior knowledge. As for NN testing, major research efforts focus on the multi-feature matching problem, i.e., multiple correct matches for a patch/feature due to multiple copy-move forgeries. The 2NN testing \cite{ref31} in classical SIFT matching cannot handle such case due to the different assumption, thus its extension versions have been designed for this purpose, such as Generalized 2NN (G2NN) \cite{ref32} and Reversed G2NN (RG2NN) \cite{ref18}, with different accuracy, speed and robustness.

For post-processing, typically, the geometric models of copy-move pairs are first evaluated from the matched keypoint set with noise (i.e., model estimation), and then the forgery regions are located in the dense-field (i.e., forgery localization). When multiple copy-move forgeries occur, the most popular global geometric model estimation algorithm, RANdom SAmple Consensus (RANSAC) \cite{ref31}, will fail. As it can only evaluate a single model, while multiple ones actually exist in such case. Therefore, separating these models is a natural idea. The existing solutions include the image-space segmentation strategy \cite{ref33} and the concept-space clustering strategy \cite{ref34}. Note that clustering strategy can better deal with multiple forgeries due to the higher discriminability, where commonly used discriminative information includes coordinates \cite{ref25,ref32,ref34}, angles \cite{ref34}, affine parameters \cite{ref20,ref35}, and offset vectors \cite{ref30}. As for localization, copy-move regions can be determined by the matched keypoint regions \cite{ref19}, or by window-wise similarity checking \cite{ref31}, e.g., Zero-mean Normalized Cross-Correlation (ZNCC) \cite{ref31} and Structural SIMilarity (SSIM) \cite{ref30}. The former has a high precision and a low recall, while the latter is just the opposite.

In summary, almost all of keypoint features are based on very shallow representations, which cannot effectively describe the local semantic properties. This leaves a fundamental flaw in the whole forensic algorithm, forcing the use of complicated matching and post-processing to remedy it.

\subsection{Deep Learning Approach}
In addition to above hand-crafted approaches, recent researchers explore the end-to-end design for copy-move forgery detection, replacing the traditional three-step pipeline.

In this approach, some popular methods are DeepNet \cite{ref45}, BusterNet \cite{ref46}, and DenseNet \cite{ref4}. Currently, on many popular benchmarks, they are typically unable to reach the detection performance w.r.t. state-of-the-art dense or sparse methods, due to the difficulty of embedding prior knowledge (especially geometric invariance) in such end-to-end networks. On the other hand, they are promising in large-scale complex forensic scenarios, because the powerful learning capability can better cope with data variations that cannot be well formalized.

\section{Proposed Method}

\subsection{Overview}
For breaking the performance bottleneck caused by semantic gap problem, we propose a BoVW model inspired copy-move forgery detection algorithm. The overall workflow is shown in Fig. 1.

\begin{figure}[!t]
	\centering
	\includegraphics[width=3.2in]{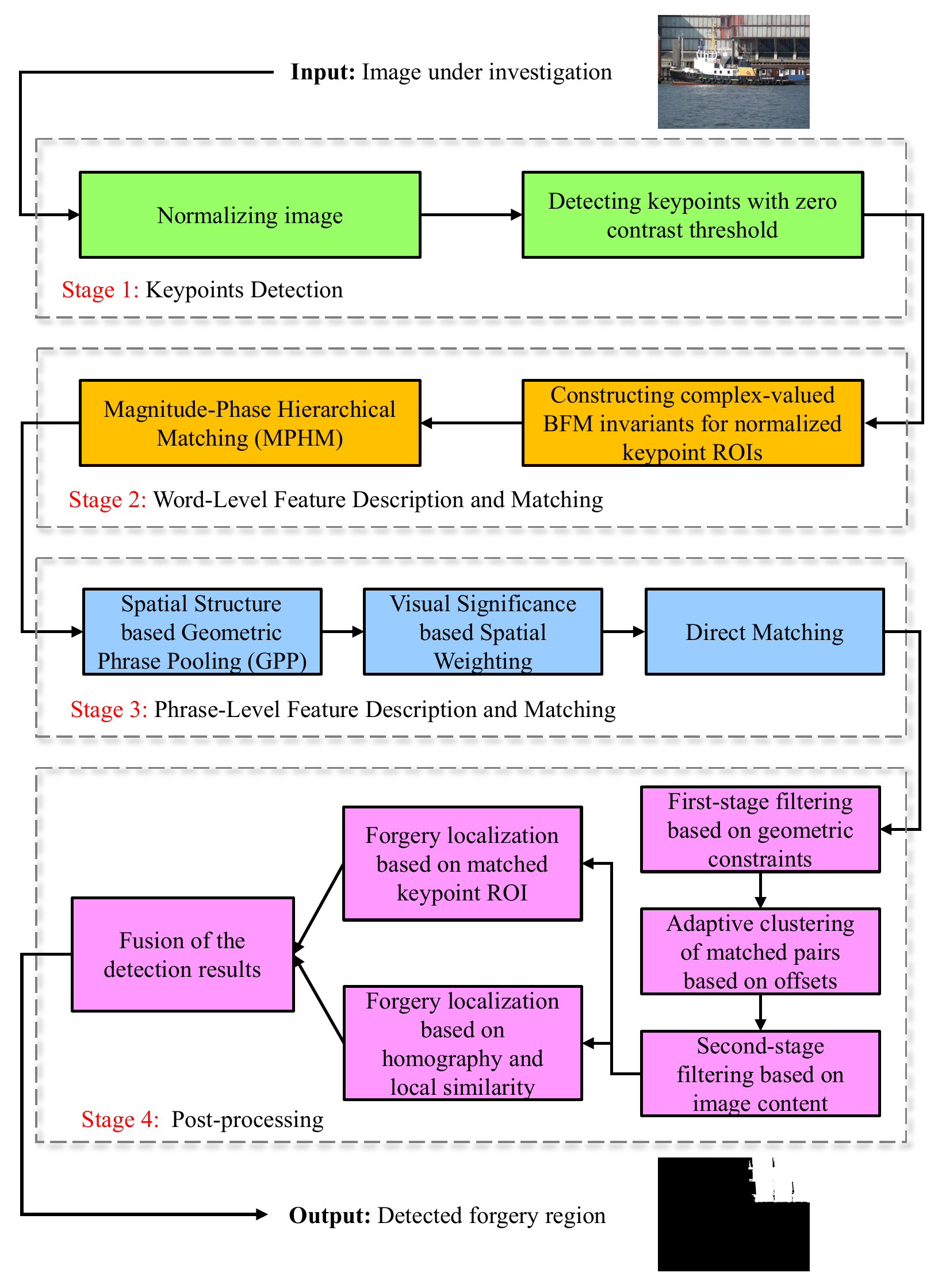}
	\caption{Illustration of the proposed copy-move forgery detection method.}

\end{figure}

In this paper, the classic three-step pipeline is generalized to a new \textbf{four-step pipeline}: keypoint detection, word-level feature description and matching, phrase-level feature description and matching, and post-processing, based on the theory of BoVW. The design aims to describe the mid-level image information via hierarchical feature extraction, and hence potentially addressing the problems related to the semantic gap.

\begin{itemize}
	\item \textbf{Keypoint detection}. We extract sufficiently dense and uniform SIFT keypoints based on two strategies: scale normalization and contrast threshold removal. They are effectiveness for detecting small/smooth copy-move regions \cite{ref30}.
	\item \textbf{Word-level feature description and matching}. Such keypoints are described by Complex-valued Moment Invariants (CMI), and then matched by magnitude-phase hierarchical matching strategy. The theoretical and experimental analyses have shown their advantages in terms of invariance and efficiency \cite{ref30}. In addition, we also propose an improved 2NN matching strategy for solving the multi-feature matching problem while reducing the time complexity. This new method is expected to replace the commonly used 2NN, G2NN and RG2NN due to the better performance on matching accuracy and efficiency.
	\item \textbf{Phrase-level feature description and matching}. The word-level matching results will be further integrated and encoded for capturing relevant information. First, the visual phrases are formed from the geometric relationships of matched keypoints, and followed by a max-pooling to construct corresponding phrase-level feature vector. Then, the features are weighted by an edge-information based heatmap, with certain foreground-background differences. Finally, such phrase-level features, associated with both spatial structure and visual saliency, will be matched as the overall matching results.
	\item \textbf{Post-processing}. The matched keypoint pairs are first filtered and clustered, and then converted into localization results on the image plane. For the filtering and clustering, two priors for correct matches are taken into account, i.e., geometric constraint and content consistency, with the ability to reduce false positives and manage multiple forgeries \cite{ref30}. For the localization, we propose a new fusion-based algorithm covering the information from geometric transformations, local similarity, and keypoint interest region. Existing works generally do not consider such rich information, and the proposed method works well in most scenarios.
\end{itemize}

\emph{Differences From the Previous Work.} Note that some techniques involved in this paper have been presented in our previous work \cite{ref30}. However, this paper is a substantial development of \cite{ref30}, w.r.t. theoretical basis, overall framework, and specific techniques. For the theoretical basis, we reveal the semantic gap problem in copy-move forgery detection, while pointing out the potential of BoVW representation model in such problem. For the overall framework, we design a novel four-step pipeline with word-to-phrase representation, as an effective practice of the above theoretical analysis. For the specific techniques, we also propose two new components, I2NN and fusion localization, which are beneficial for further improving the performance w.r.t. multiple forgeries and similar-but-genuine backgrounds. None of the above three ingredients are covered in previous \cite{ref30}, also in the field of copy-move forgery detection to the best of our knowledge.

The following Sections \uppercase\expandafter{\romannumeral3}-B to \uppercase\expandafter{\romannumeral3}-E will introduce the above components respectively. Considering the focus of this paper, the techniques inherited from our previous work \cite{ref30}, mainly in Stages 2 and 4, will not be described in detail.

\subsection{Keypoint Detection}

Similar to many existing copy-move forgery detection algorithms, the classical SIFT \cite{ref31} is adopted for detecting geometric invariant keypoints with corresponding interest regions. However, the general setting of SIFT cannot guarantee that sufficient keypoints are extracted in small or smooth regions, which in turn leads to the failure in detecting copy-move forgery for such regions. To address this challenge, two tricks have been explored in recent work \cite{ref19}, and they will also be used in our algorithm.

The first trick enhances the keypoint detection performance in smooth regions by \textbf{removing the contrast threshold}. In many classical computer vision applications, such as image registration, ensuring a high repeatability of keypoints is a primary goal, rather than the coverage. For this reason, there is a step in SIFT to eliminate the unstable keypoints, i.e., passing only the keypoints with sufficiently high local contrast. However, for copy-move detection, keypoint coverage of the forgery regions plays an even more important role than the repeatability. Therefore, we skip such a step and directly set the contrast threshold ${T_{con}} = 0$ in the implementation.

The second trick enhances the keypoint detection performance in small regions by \textbf{increasing the image resolution}. When the copy-move regions contain few pixels, it has difficulties in achieving high keypoint coverage on such regions, even though the contrast threshold ${T_{con}}$ is set to 0. The simplest solution is to increase the resolution of the whole image, and thus also increase the number of pixels in the potentially forgery region, allowing more candidate keypoints to be detected. However, it is clear that this strategy requires a trade-off between coverage and efficiency. In the implementation of Li et al. \cite{ref19}, the scale factor   is set to a fixed value ($s = 2$), resulting in expensive calculations for high-resolution images. For this reason, in our previous work \cite{ref30}, $s$ is a variable value and is determined according to the resolution of the input image. Although coverage and efficiency are basically balanced, the image size may changes over a wide interval, and some fixed parameters cannot adapt to it. In this paper, we refine this strategy and the scaling factor $s$ defined as:
\begin{equation}
	s = \left\{ {\begin{array}{*{20}{c}}
			{\frac{{\max ({N_I},{M_I})}}{{3000}}},&{\max ({N_I},{M_I}) < 3000}\\
			1,&{\max ({N_I},{M_I}) \ge 3000}
	\end{array}} \right.,
\end{equation}
where ${N_I} \times {M_I}$ is the resolution for the input image $I$. The (1) normalizes the long edge of the low-resolution (i.e., $\max ({N_I},{M_I}) < 3000$) image to 3000 pixels, without changing the aspect ratio. The design aims to extract sufficient keypoints at a reasonable time cost, while reducing the sensitivity of parameters/thresholds to the resolution for the following stages. Here, the up-sampling is implemented by bicubic interpolation. The reason why a similar normalization process is not performed on high-resolution (i.e., $\max ({N_I},{M_I}) \ge 3000$) images is that down-sampling will inevitably produce a certain degree of information loss, which may be more harmful to feature description and post-processing.

Through the above strategies, one can obtain a set of SIFT keypoints, ${\bf{{K}}} = \{ {{\bf{k}}_k} = {(x,y,\sigma )_k}\} _{k = 1}^{\#_K}$, where $(x,y)$ is the coordinate, $\sigma $ is the scale and $\#_K$ is the number of keypoints. For better visualization, two examples of the keypoint detection are shown in Fig. 2. As can be seen that, after the optimization based on two strategies, the keypoints are more uniformly and densely distributed over the image plane, ensuring the coverage of the copy-move regions.
\begin{figure}[!t]
	\centering
	\subfigure[]{\includegraphics[width=1.6in]{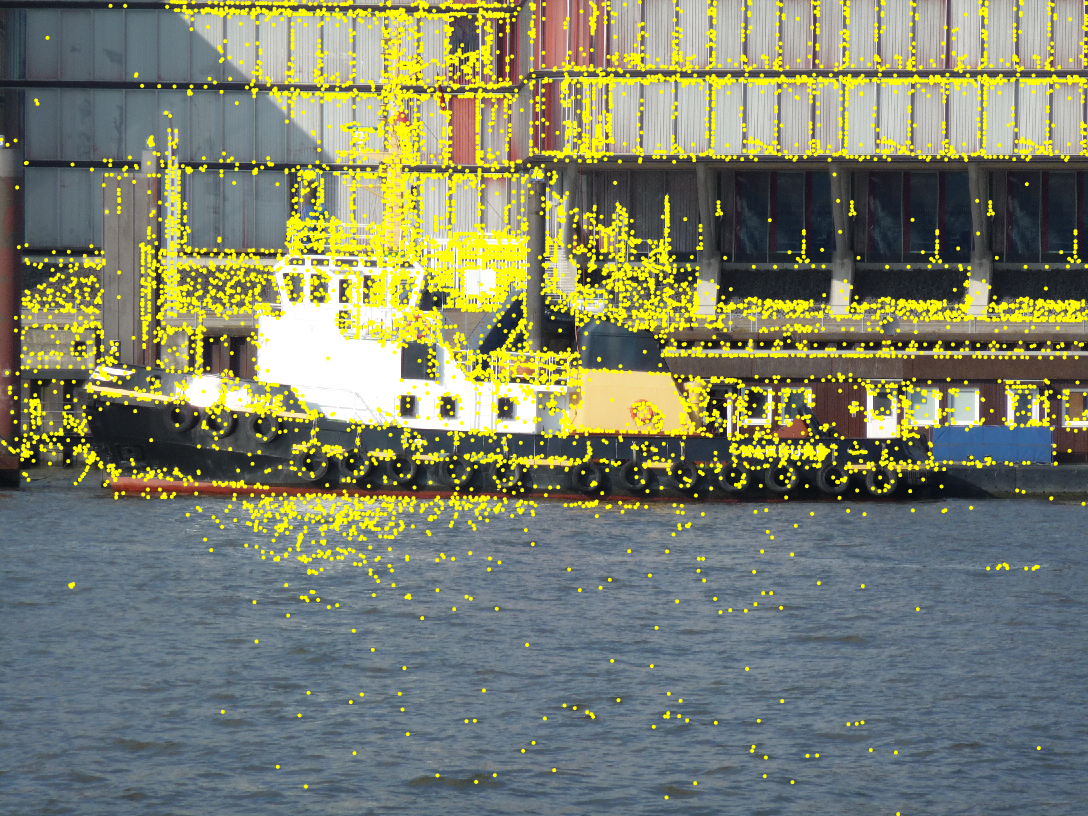}}
	\subfigure[]{\includegraphics[width=1.6in]{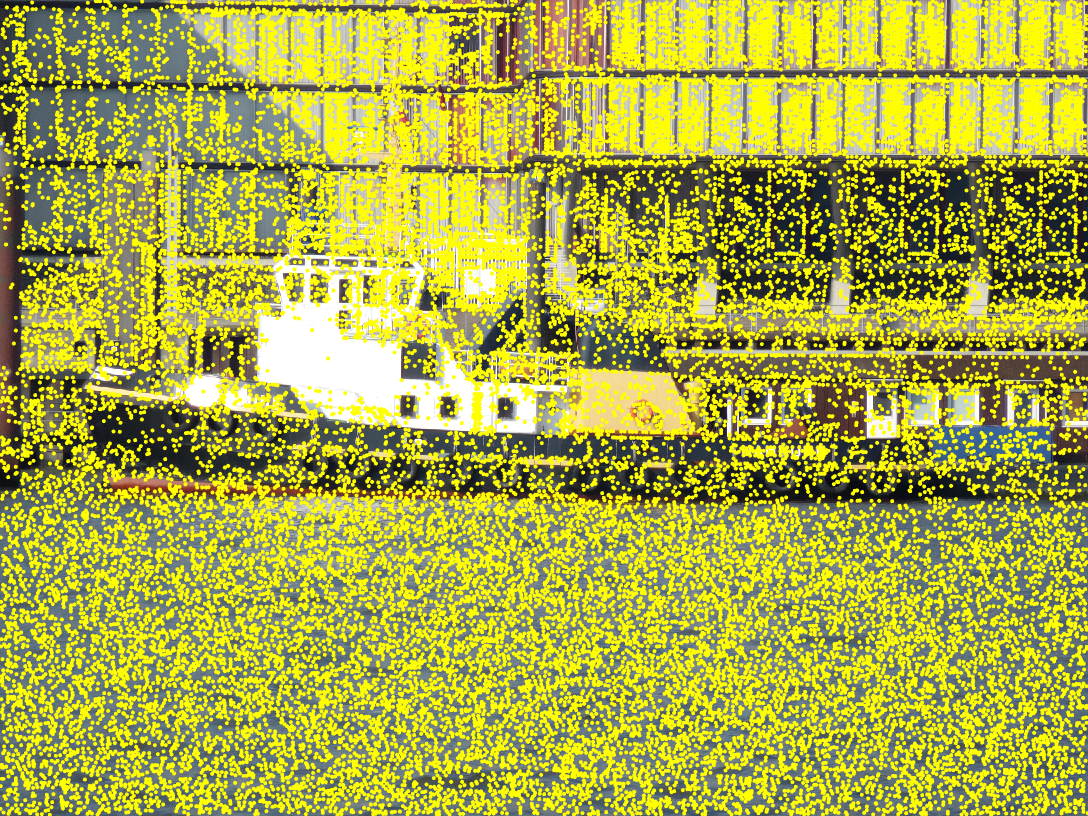}}
	\centering
	\caption{Keypoints detection: (a) the SIFT under common setting and (b) the SIFT by removing the contrast threshold and increasing the image resolution.}
\end{figure}

\subsection{Word-Level Feature Description and Matching}
In this section, the SIFT keypoints will be roughly described and matched at the visual-word level. For this task, we continue to use the complex-valued moment invariant description and magnitude-phase hierarchical matching in our previous work \cite{ref30}, due to their advantages in representation ability and matching speed. Here, we will propose a new NN testing strategy to replace the 2NN in the matching, considering its difficulty in handling multiple copy-move forgeries.

Through the description, one can obtain a set of features ${\bf{F}} = \{ {{\bf{f}}_k}\} _{k = 1}^{\#_K} $, which corresponds to the keypoint set ${{\bf{{K}}} = \{ {{\bf{k}}_k}\} _{k = 1}^{\#_K}}$. Based on a specific distance metric, the matching starts with a search for several closest features to a given feature, i.e., \emph{NN searching}. Mathematically, for a feature ${{\bf{f}}_{(0)}}$ (w.r.t. keypoint ${{\bf{k}}_{(0)}}$), its first $n$ nearest neighbors in the feature set ${\bf{F}}$ are denoted as $\{ {{\bf{f}}_{(1)}},{{\bf{f}}_{(2)}},...,{{\bf{f}}_{(n)}}\} $ (w.r.t. keypoints $\{ {{\bf{k}}_{(1)}},{{\bf{k}}_{(2)}},...,{{\bf{k}}_{(n)}}\} $), i.e., the NN set of ${{\bf{f}}_{(0)}}$; the distances between ${{\bf{f}}_{(0)}}$ and $\{ {{\bf{f}}_{(1)}},{{\bf{f}}_{(2)}},...,{{\bf{f}}_{(n)}}\} $ are denoted as $\{ {d_1},{d_2},...,{d_n}\} $. Based on a certain prior, the final judgment of whether ${{\bf{f}}_{(0)}}$ and $\{ {{\bf{f}}_{(1)}},{{\bf{f}}_{(2)}},...,{{\bf{f}}_{(n)}}\} $ are matched pairs is formed, i.e., \emph{NN testing}. In this regard, the existing NN testing strategies are mainly based on distance information in $\{ {d_1},{d_2},...,{d_n}\} $, including two types \cite{ref36}:

\begin{itemize}
	\item \textbf{Absolute distance based NN testing}. It simply regards ${{\bf{f}}_{(0)}}$ (w.r.t. ${{\bf{k}}_{(0)}}$) and ${{\bf{f}}_{(1)}}$ (w.r.t. ${{\bf{k}}_{(1)}}$) as a matched pair when ${d_1}$ is less than a predefined threshold. In the complex tasks (e.g., image registration), however, the distance between a feature and its counterpart varies widely, due to diversity of imaging conditions (e.g., illumination and viewing angle). Therefore, such method is usually unsatisfactory in matching accuracy \cite{ref36}.
	\item \textbf{Relative distance based NN testing}. This path is generally based on the ratio of the distances, which exhibits a higher tolerance for imaging conditions, and thus serving as a common choice \cite{ref36}
\end{itemize}

In the field of copy-move forgery detection, relative distance based NN testing methods mainly include 2NN \cite{ref31}, G2NN \cite{ref32}, and RG2NN \cite{ref18}.

\begin{itemize}
	\item \emph{2NN}. For given ${{\bf{f}}_{(0)}}$, this strategy only searches the first two nearest neighbors, i.e., $\{ {{\bf{f}}_{(1)}},{{\bf{f}}_{(2)}}\} $ with $\{ {d_1},{d_2}\} $. It regards ${{\bf{f}}_{(0)}}$ (w.r.t. ${{\bf{k}}_{(0)}}$) and ${{\bf{f}}_{(1)}}$ (w.r.t. ${{\bf{k}}_{(1)}}$) as a matched pair when ${d_1}/{d_2}$ is less than a predefined threshold. It should be noted that the 2NN is based on the following assumption: there is at most one true match (semantically) for a keypoint. This assumption is in line with some computer vision tasks, but not applicable to copy-move detection due to the existence of multiple forgeries. For such scenarios, ${d_1}/{d_2} \approx 1$ since both ${{\bf{f}}_{(1)}}$ and ${{\bf{f}}_{(2)}}$ are true matches.
	\item \emph{G2NN}. It is a generalized version of 2NN and is especially designed for multiple copy-move forgeries detection. By the order $i = 1,2,...,n - 1$, it calculates ${d_i}/{d_{i + 1}}$ separately, until a position $k$ allowing ${d_k}/{d_{k + 1}}$ greater than a predefined threshold; then the ${{\bf{f}}_{(0)}}$ and $\{ {{\bf{k}}_{(1)}},...,{{\bf{k}}_{(k)}}\} $ are considered as matched pairs. However, in the same way, when multiple forgeries exists with ${d_1}/{d_2} \approx 1$, such a loop will terminate prematurely.
	\item \emph{RG2NN}. It is a reversed version of G2NN for solving the above premature termination problem. By the order $i = n,n - 1,...,2$, it calculates ${d_{i - 1}}/{d_i}$ separately, until a position $k$ allowing ${d_{k - 1}}/{d_k}$ less than a predefined threshold; then the ${{\bf{f}}_{(0)}}$ and $\{ {{\bf{k}}_{(1)}},...,{{\bf{k}}_{(k - 1)}}\} $ are considered as matched pairs. Compared with G2NN, RG2NN is theoretically more suitable for multi-feature matching. However, since $k$ is generally much less than $n$, such NN testing often requires a large computational cost.
\end{itemize}

Fig. 3 gives an illustration of 2NN, G2NN and RG2NN. Here, for given keypoint ${{\bf{k}}_{(0)}}$, its first NN ${{\bf{k}}_{m1}}$ and second NN ${{\bf{k}}_{m2}}$ are true matches, the remaining $n - 2$ neighbors $\{ {{\bf{k}}_{u1}},{{\bf{k}}_{u2}},...,{{\bf{k}}_{u(n - 2)}}\} $ are irrelevant keypoints, and the feature distances between ${{\bf{k}}_{(0)}}$ and $\{ {{\bf{k}}_{m1}},{{\bf{k}}_{m2}},{{\bf{k}}_{u1}},{{\bf{k}}_{u2}},...,{{\bf{k}}_{u(n - 2)}}\} $ are $\{ {d_1},{d_2},...,{d_n}\} $, respectively. Since ${d_1}/{d_2} \approx 1$, the 2NN and G2NN directly treat ${{\bf{k}}_{m1}}$ and ${{\bf{k}}_{m2}}$ as irrelevant. The RG2NN is able to match ${{\bf{k}}_{m1}}$ and ${{\bf{k}}_{m2}}$, but a total of $n - 2$ tests for a keypoint results in a higher complexity.

\begin{figure}[!t]
	\centering
	\includegraphics[width=2in]{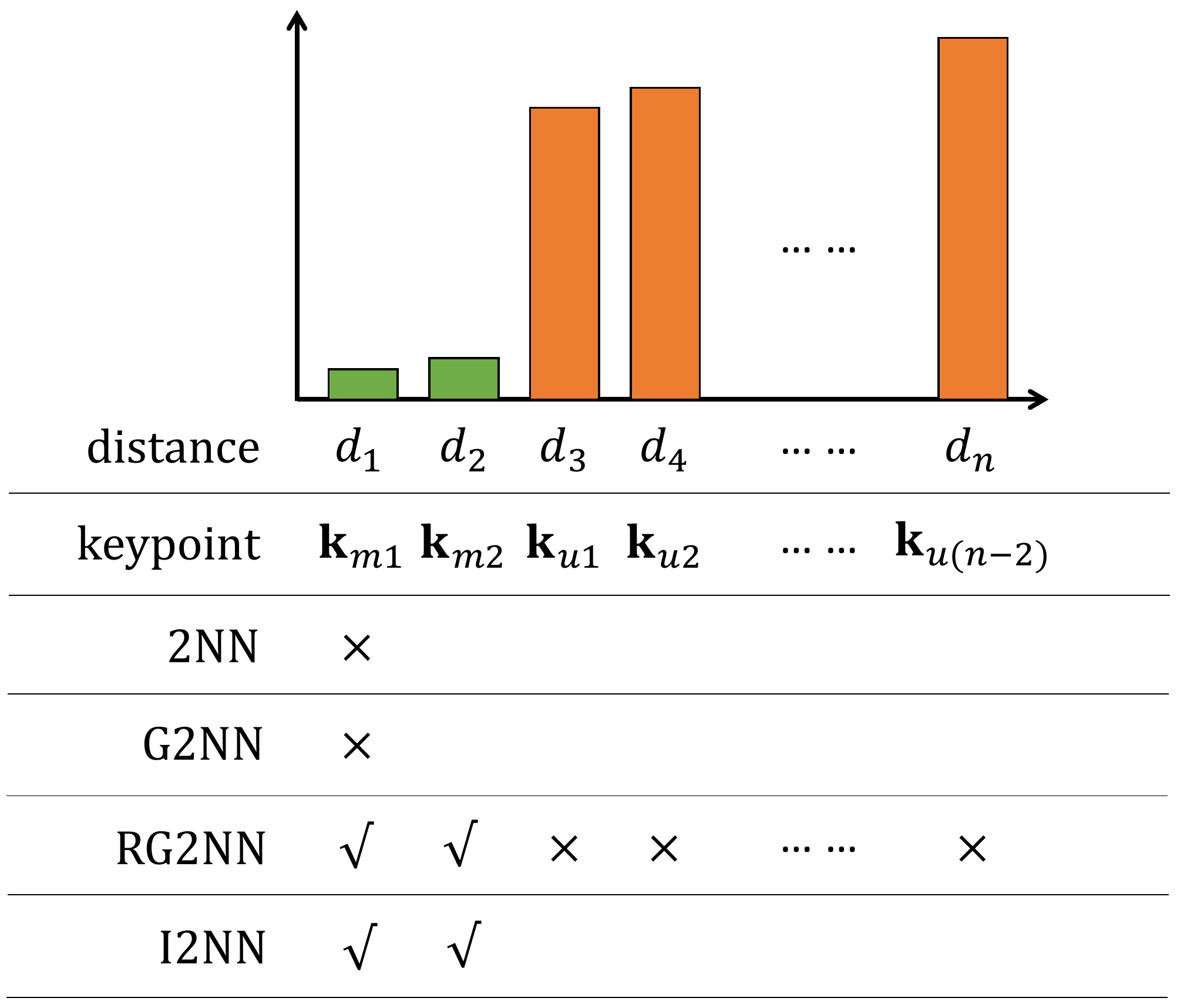}
	\caption{Illustration of the different keypoint matching strategies.}
\end{figure}

To handle multiple copy-move forgeries detection with less complexity, we propose an \textbf{Improved 2NN (I2NN)} algorithm. Specifically, for the given ${{\bf{f}}_{(0)}}$ (w.r.t. ${{\bf{k}}_{(0)}}$) and its first two nearest neighbors $\{ {{\bf{f}}_{(1)}},{{\bf{f}}_{(2)}}\} $ (w.r.t. $\{ {{\bf{k}}_{(1)}},{{\bf{k}}_{(2)}}\} $), when the following single-feature matching conditions,
\begin{equation}
	\begin{split}
		&\frac{d_1}{d_2} \le {T_{sml{\rm{ - }}rat}}, \\
		&{\left\| {{{\bf{k}}_{(0)}} - {{\bf{k}}_{(1)}}} \right\|_2} \ge {T_{dist}},
	\end{split}
\end{equation}
are satisfied, the features ${{\bf{f}}_{(0)}}$ and ${{\bf{f}}_{(1)}}$ (w.r.t. keypoints ${{\bf{k}}_{(0)}}$ and ${{\bf{k}}_{(1)}}$) are considered as a matched pair; or when the following multi-feature matching conditions,
\begin{equation}
	\begin{split}
	& d_2 \le {T_{sml{\rm{ - }}abs}},\\
	&{\left\| {{{\bf{k}}_{(0)}} - {{\bf{k}}_{(1)}}} \right\|_2},{\left\| {{{\bf{k}}_{(0)}} - {{\bf{k}}_{(2)}}} \right\|_2} \ge {T_{dist}},
\end{split}
\end{equation}
are satisfied, the features ${{\bf{f}}_{(0)}}$ and $\{ {{\bf{f}}_{(1)}},{{\bf{f}}_{(2)}}\} $ (w.r.t. keypoints ${{\bf{k}}_{(0)}}$ and $\{ {{\bf{k}}_{(1)}},{{\bf{k}}_{(2)}}\} $) are all considered as matched pairs, where ${T_{sml{\rm{ - }}rat}}$ and ${T_{sml{\rm{ - }}abs}}$ are the relative and absolute thresholds for measuring the distances between the features (${T_{sml{\rm{ - }}rat}} = 0.6$ and ${T_{sml{\rm{ - }}abs}} = 0.1$ in our implementation), and ${T_{dist}}$ is a threshold for measuring the distances between the keypoint positions (${T_{dist}} = 50$ in our implementation). Note that ${d_2} = {\left\| {{{\bf{f}}_{(0)}} - {{\bf{f}}_{(2)}}} \right\|_2}$ used in (3), instead of ${d_1} = {\left\| {{{\bf{f}}_{(0)}} - {{\bf{f}}_{(1)}}} \right\|_2}$, since ${d_2}$ is only small when multiple true matches exist.

Fig. 3 shows the comparison of I2NN and existing strategies. The proposed I2NN considers both absolute and relative distances, and thus achieves a limited ability for multi-feature matching, which draws a distinction between 2NN and G2NN. Moreover, the I2NN checks only the first two nearest neighbors, with less time complexity than RG2NN. Note that, as far as copy-move detection task is concerned, we actually do not need the full multi-feature matching ability like RG2NN. Since the number of the copy-move regions is rarely large, the ability of I2NN to match two nearest neighbors is generally sufficient in practice. As an illustration, the matching results of 2NN, G2NN, RG2NN, and I2NN for real image are given in Fig. 4. It can be seen that the proposed I2NN is able to deal with multiple copy-move forgeries, with much fewer mismatches and complexity.
\begin{figure}[!t]
	\centering
	\subfigure[2NN]{\includegraphics[width=1.4in]{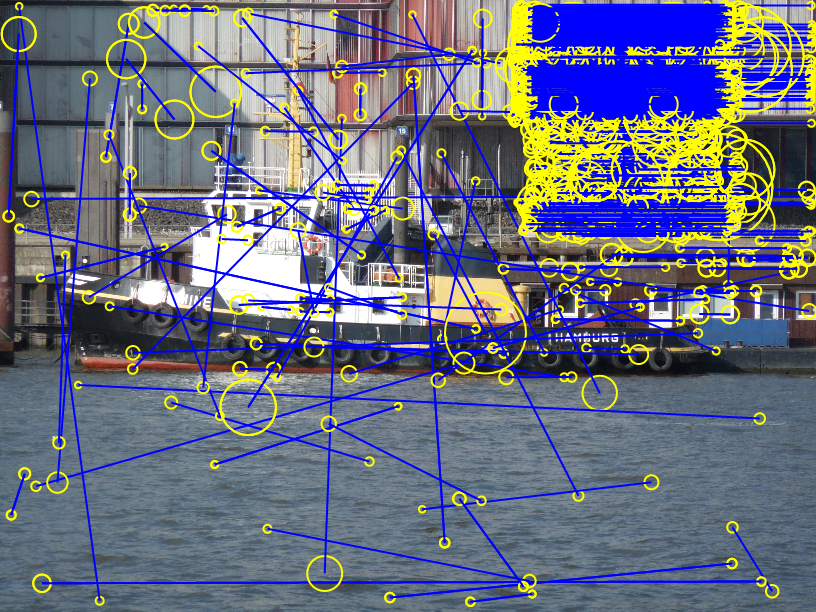}}
	\subfigure[G2NN]{\includegraphics[width=1.4in]{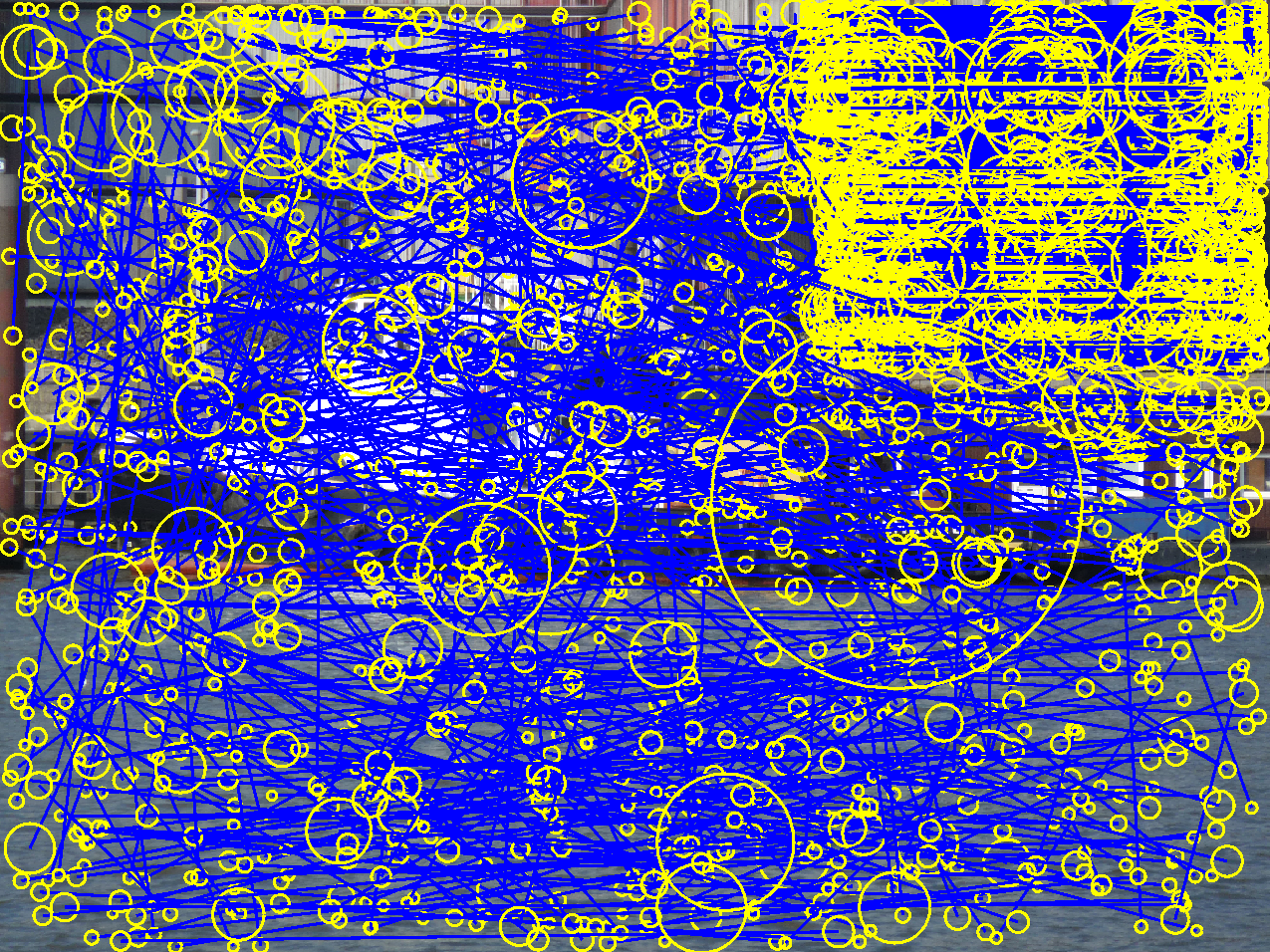}}
	\subfigure[RG2NN]{\includegraphics[width=1.4in]{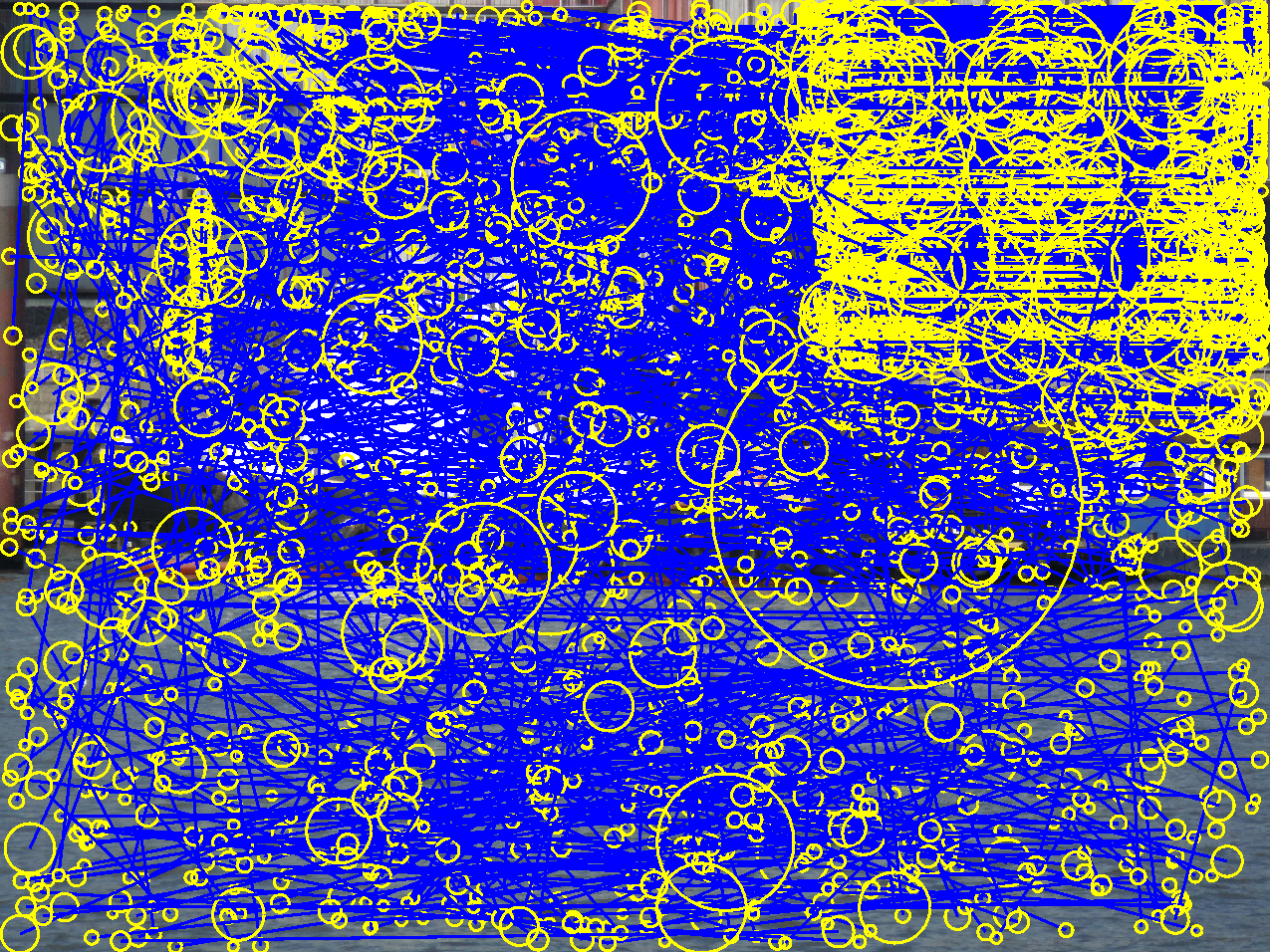}}
	\subfigure[I2NN]{\includegraphics[width=1.4in]{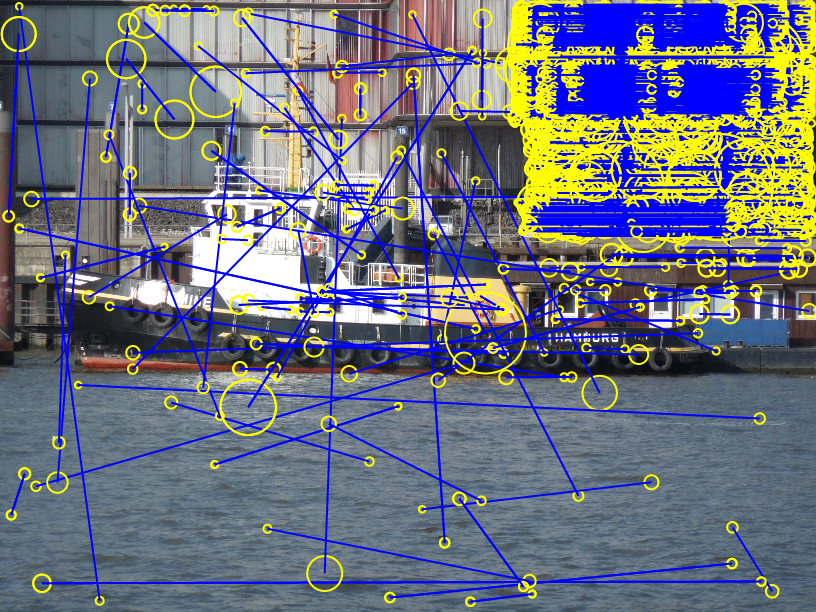}}
	
	\centering
	\caption{Word-level feature matching by different methods.}
\end{figure}

\subsection{Phrase-Level Feature Description and Matching}
In this section, the above matching results will be further described and matched at the visual-phrase level. Existing methods generally feed the word-level results directly to post-processing, where the semantic gap inevitably leads to unsatisfactory forgery localization results. Therefore, we recommend performing a hierarchical feature extraction and matching process to capture more semantic properties. Specifically, the proposed phrase-level feature description and matching algorithm includes the following steps:
\begin{itemize}
\item \textbf{Step 1:} Geometric phrase pooling based on spatial structure;
\item \textbf{Step 2:} Spatial weighting based on visual saliency;
\item \textbf{Step 3:} Direct matching of phrase-level features.
\end{itemize}

\subsubsection{Geometric Phrase Pooling based on Spatial Structure}

This step uses the geometric relationships of keypoints to form visual phrases, and then performs max-pooling to obtain the representation reflecting spatial structure. The main purpose is avoiding the \emph{isolation} of the local descriptors in existing copy-move detection works, where the features are processed very independently, ignoring their relationships in the image plane. We design a mid-level connection between low-level descriptors and high-level concepts for the copy-move detection scenarios, inspired by the work of Xie et al. \cite{ref37}.

Going back to Sections \uppercase\expandafter{\romannumeral3}-C, with keypoint set ${\bf{{K}}} = \{ {{\bf{k}}_k}\} _{k = 1}^{\#_K}$, one can obtain a set of word-level features ${{\bf{F}}^w} = \{ {\bf{f}}_k^w\} _{k = 1}^{\#_K}$ and a set of word-level matched pairs ${{\bf{P}}^w} = \{ {\bf{p}}_k^w = {{\rm{(}}{\bf{k}},{\bf{k'}}{\rm{)}}_k}\} _{k = 1}^{\#_{P@W}}$, where $\#_{P@W}$ is the number of pairs at the word level; the corresponding set of word-level matched keypoints denotes as ${{\bf{{ K}}}^{w}} = \{ {{\bf{k}}_k}\} _{k = 1}^{2\#_{P@W}} \in {\bf{K}}$. Based on the above notations, a set of \textbf{visual words} can be defined:
\begin{equation}
	{\bf{D}} = \{ ({{\bf{k}}_1},{\bf{f}}_1^w),({{\bf{k}}_2},{\bf{f}}_2^w),...,({{\bf{k}}_{2\#_{P@W}}},{\bf{f}}_{2\#_{P@W}}^w)\} ,
\end{equation}
where visual word $({{\bf{k}}_k},{\bf{f}}_k^w)$ is the combination of keypoint ${{\bf{k}}_k}$ and corresponding word-level feature ${\bf{f}}_k^w$. For a visual word $({{\bf{k}}_c},{\bf{f}}_c^w)$ in ${\bf{D}}$, its \textbf{visual phrase} is defined as:
\begin{equation}
	{\bf{G}} = \{ ({{\bf{k}}_c},{\bf{f}}_c^w),({{\bf{k}}_{s1}},{\bf{f}}_{s1}^w),({{\bf{k}}_{s2}},{\bf{f}}_{s2}^w),...,({{\bf{k}}_{sK}},{\bf{f}}_{sK}^w)\},
\end{equation}
where word $({{\bf{k}}_c},{\bf{f}}_c^w)$ is named as central word of phrase ${\bf{G}}$, resting $\{ ({{\bf{k}}_{s1}},{\bf{f}}_{s1}^w),({{\bf{k}}_{s2}},{\bf{f}}_{s2}^w),...,({{\bf{k}}_{sK}},{\bf{f}}_{sK}^w)\} $ are the $K$ visual words closest to the center word $({{\bf{k}}_c},{\bf{f}}_c^w)$ in the image plane, called side words of phrase ${\bf{G}}$ ($K = 3$ in our implementation). To speed up the search for side words, we build a KD-tree using the keypoint location information from word set ${\bf{D}}$, based on the VLFEAT package \cite{ref38}.

Suppose the dimension of word-level feature ${\bf{f}}_k^w$ is $d$, written as ${\bf{f}}_k^w = \{ f_{k,1}^w,f_{k,2}^w,...,f_{k,d}^w\} $. Based on the geometric phrase ${\bf{G}}$, the corresponding \textbf{geometric phrase feature} ${\bf{f}}_c^g$ for the keypoint ${{\bf{k}}_c}$ can be obtained through \textbf{geometric phrase pooling}:
\begin{equation}
	\begin{split}
		{\bf{f}}_c^g & = \{ f_{c,1}^g,f_{c,2}^g,...,f_{c,d}^g\} \\
		& = \{ f_{c,1}^w + \max (f_{sk,1}^w)_{k = 1}^K,f_{c,2}^w  + \max (f_{sk,2}^w)_{k = 1}^K, \\
		& ...,f_{c,d}^w + \max (f_{sk,d}^w)_{k = 1}^K\} .
	\end{split}
\end{equation}

The intuitive interpretation of (6) is to first find the maximum value among the side words $\{ {\bf{f}}_{s1}^w,{\bf{f}}_{s2}^w,...,{\bf{f}}_{sK}^w\} $ along each dimension of the feature, and then add the result to the corresponding dimension of the center word feature ${\bf{f}}_c^w$, finally forming a phrase-level feature ${\bf{f}}_c^g$. For more details on geometric phrase pooling, please refer to \cite{ref37}.

The generated phrase-level features are with following properties: 1) ${\bf{f}}_c^g$ maintains the same dimensions as ${\bf{f}}_c^w$ and thus does not increase the matching complexity; 2) ${\bf{f}}_c^g$ combines the complete information of the central word ${\bf{f}}_c^w$ with the partial information of the adjacent words $\{ {\bf{f}}_{s1}^w,{\bf{f}}_{s2}^w,...,{\bf{f}}_{sK}^w\} $, exhibiting better discriminability than word-level representation; and 3) the max-pooling allows an order-independent information extraction for $\{ {\bf{f}}_{s1}^w,{\bf{f}}_{s2}^w,...,{\bf{f}}_{sK}^w\} $, meaning the invariance w.r.t. geometric changes, especially rotation.

Fig. 5 shows an intuition of the geometric phrase pooling operation in copy-move detection. Here, the central words of geometric phrases ${{\bf{G}}_1}$, ${{\bf{G}}_2}$, and ${{\bf{G}}_3}$ have similar features, so they are regarded as matched pairs in the word-level process. Through geometric phrase pooling, such isolated features are upgraded to phrase-level features with spatial structure, and thus become more informative. For the case in Fig. 5, the correct matching between ${{\bf{G}}_1}$ and ${{\bf{G}}_2}$ is preserved in phrase-level process, due to the similarity in both center words and side words (even with different orders). As for ${{\bf{G}}_1}$ and ${{\bf{G}}_3}$, this mismatched pair is passed in the word-level process, but is eliminated in the phrase-level process by the inconsistency of side words.
\begin{figure}[!t]
	\centering
	\includegraphics[width=2.2in]{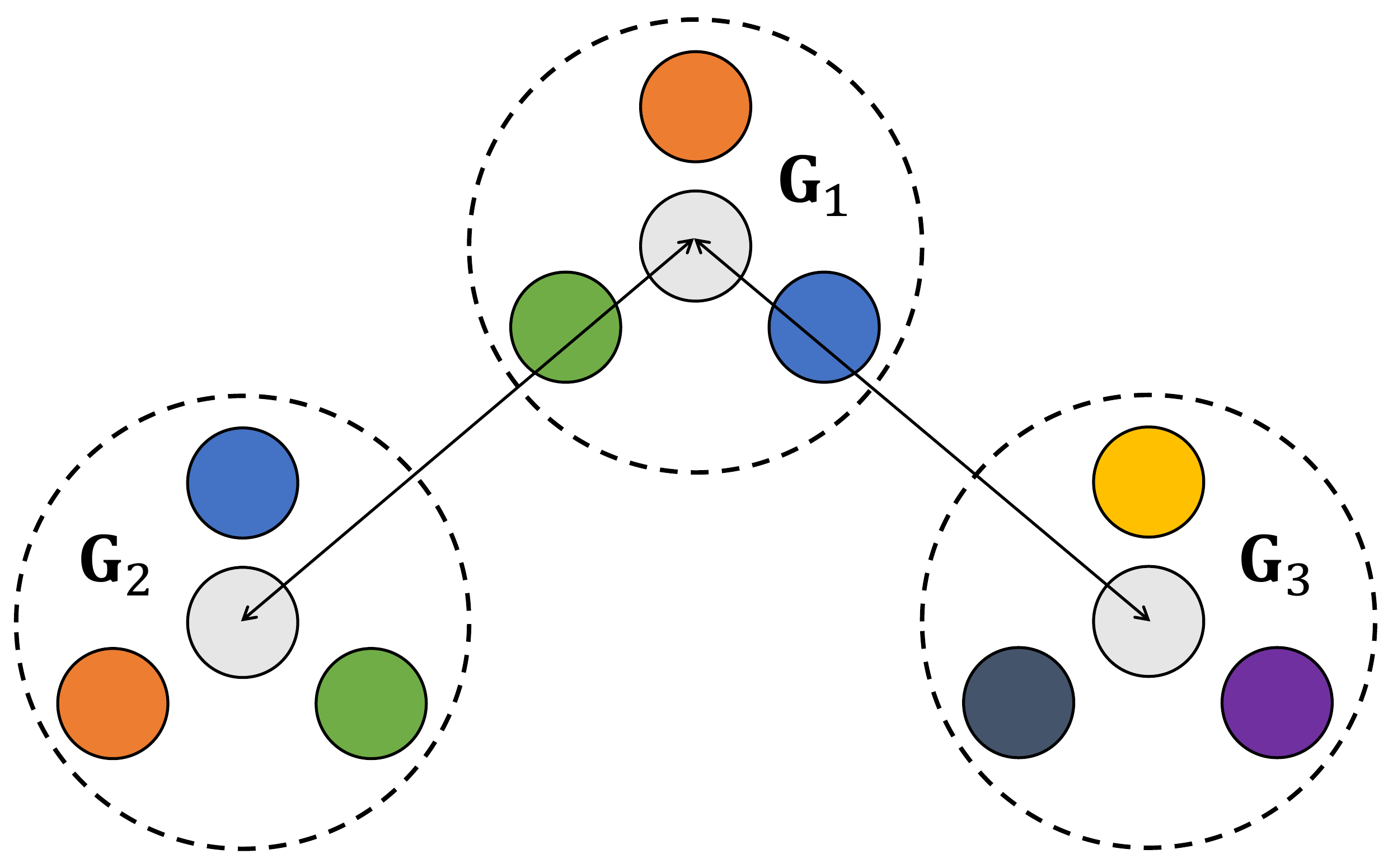}
	\caption{Illustration of the geometric phrase pooling.}
\end{figure}

\subsubsection{Spatial Weighting based on Visual Saliency}
This step uses the edge information of image to form a heat map, and then weights the phrase-level features to strengthen the difference in visual saliency. The main purpose is avoiding the \emph{equality} of the local descriptors in existing copy-move detection works, where the features are processed very equivalently, ignoring their visually differences in the image plane. We design a visual saliency based spatial weighting for the copy-move detection scenarios, with a corresponding acceleration method, inspired by the work of Xie et al. \cite{ref37}.

According to the relevant research on visual perception, the human visual system is more sensitive to the edge of object, which is the basis for the semantic interpretation of the shapes. Therefore, it is reasonable to generate a saliency map through edge information. For a gray image $\{ f(i,j):(i,j) \in [1,...,{N_I}] \times [1,...,{M_I}]\} $, its \textbf{edge map} is denoted as:
\begin{equation}
	\{ e(i,j):(i,j) \in [1,...,{N_I}] \times [1,...,{M_I}]\},
\end{equation}
where $e(i,j) \in [0,1]$ is the edge response/possibility at the position $(i,j)$. For generating (7), a  edge detection algorithm by Zitnick et al. \cite{ref39} is adopted, considering its efficient calculation and sufficiently accurate results. Based on the edge map $e$, the \textbf{saliency heat map} $\{ h(i,j):(i,j) \in [1,...,{N_I}] \times [1,...,{M_I}]\} $ is defined by accumulating the edge responses with Gaussian blurring:
\begin{equation}
	\begin{split}
	h(i,j) & = {e} \otimes {G_{ij}}(i',j') \\
	& = \sum\limits_{i' = 1}^{{N_I}} {\sum\limits_{j' = 1}^{{M_I}} {e(i',j')\exp ( - {T_\sigma } \cdot ||(i,j) - (i',j')|{|_2})} } ,
	\end{split}
\end{equation}
where $h(i,j)$ is the visual saliency of image at $(i,j)$; ${G_{ij}}(i',j') = \exp ( - {T_\sigma } \cdot ||(i,j) - (i',j')|{|_2})$ is the Gaussian kernel function centered at $(i,j)$; ${T_\sigma }$ is the smoothing threshold (${T_\sigma } = {\rm{0}}{\rm{.001}}$ in our implementation). Here, the smaller the ${T_\sigma }$, the larger the influence range of a specific response in the heat map.

In practice, the (8) needs to enumerate all $(i',j')$ (a total of ${N_I} \times {M_I}$) for each position $(i,j)$, resulting in a very high complexity $O({N_I}^2{M_I}^2)$. For this, an approximate formula is given by ignoring the position $(i',j')$ with lower ${G_{ij}}(i',j')$. Specifically, based on the \textbf{three-sigma rule} \cite{ref48}, the enumeration range is restricted to a square region $S$ with center $(i,j)$ and length $L = \left\lceil {{3 \mathord{\left/
			{\vphantom {3 {\sqrt {2{T_\sigma }} }}} \right.
			\kern-\nulldelimiterspace} {\sqrt {2{T_\sigma }} }}} \right\rceil $, without significant loss of accuracy:
\begin{equation}
	\begin{split}
	& (i',j') \in S = \left[ {i - L,i + L} \right] \times \left[ {j - L,j + L} \right].
	\end{split}
\end{equation}

Then, the approximate version of (8) can be derived:
\begin{equation}
	h(i,j) \approx \sum\limits_{(i',j') \in S} {e(i',j')\exp ( - {T_\sigma } \cdot ||(i,j) - (i',j')|{|_2})} ,
\end{equation}
where the complexity is reduced to $O(4{N_I}{M_I}{L^2})$. This is acceptable for low-resolution scenarios like in \cite{ref37}, but not for copy-move detection, due to the large ${N_I} \times {M_I}$ and $L$. An equivalent but efficient implementation of (10) can be achieved by the help of \textbf{Fast Fourier Transform (FFT)} and \textbf{Convolution Theorem}:
\begin{equation}
	{h} \approx {\mathcal{F}^{ - 1}}(\mathcal{F}(e)\mathcal{F}({G_{L,L}}(i',j'))) ,
	(i',j') \in S.
\end{equation}
with complexity $O(3{N_I}{M_I}\log ({N_I}{M_I}) + 4{N_I}{M_I})$.

Based on the geometric phrase feature ${\bf{f}}_c^g$ and the saliency heat map $h$, the\textbf{ weighted geometric phrase feature} ${\bf{f}}_c^p$ corresponding to the keypoint ${{\bf{k}}_c}$ can be defined as:
\begin{equation}
	{\bf{f}}_c^p = h({{\bf{k}}_c}) \cdot {\bf{f}}_c^g .
\end{equation}

Note that our motivation is to enhance the discriminability of the features, rather than restricting the detection to salient regions. Specifically, the copy and move regions will share equal (almost equal) weights, and hence, in fact, the small-but-equal weights (e.g. in copy-moved backgrounds) will not affect the correct matching. In addition, the weights in our implementation range from 1 to 2, rather than the common 0 to 1.

Fig. 6 shows the examples of edge map and saliency heat map. Here, the extracted edges are semantically relevant, and the resulting heat map well reflects the visual saliency differences in different regions. It is worth noting that the corresponding positions in the copy-move regions have similar weights, while the weights in other regions are more diverse. This fact clearly facilitates the further enhancement of phrase-level features in terms of discriminability.

\begin{figure}[!t]
	\centering
	\subfigure[Edge map]{\includegraphics[width=1.6in]{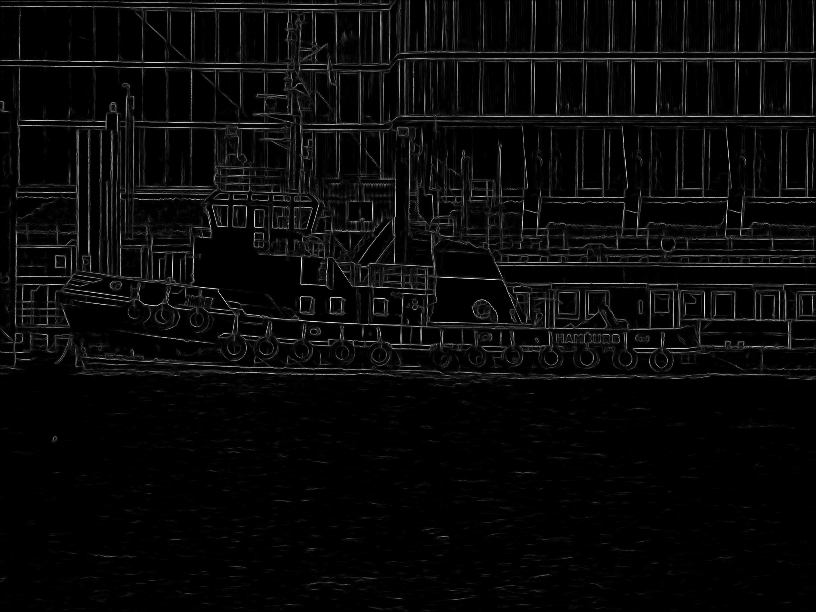}}
	\subfigure[Saliency heat map]{\includegraphics[width=1.6in]{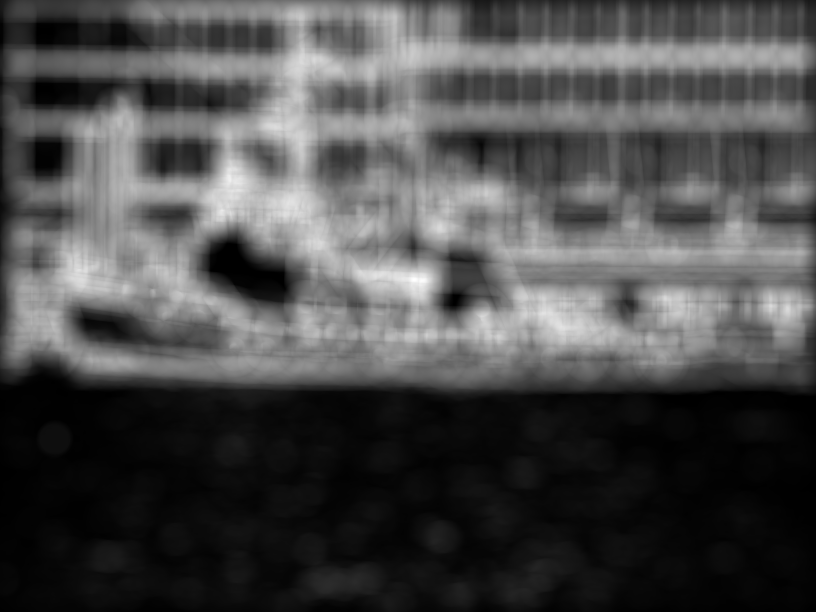}}
	
	\centering
	\caption{Edge-based spatial weighting.}
\end{figure}

\subsubsection{Direct Matching of Phrase-Level Features}

In this step, the obtained phrase-level features reflecting the spatial structure relationships and visual saliency differences will be directly matched as the final matching result. For the word-level matched keypoint set ${{\bf{{ K}}}^{w}} = \{ {{\bf{k}}_k}\} _{k = 1}^{2\#_{P@W}}$ and the corresponding phrase-level feature set ${{\bf{F}}^p} = \{ {\bf{f}}_k^p\} _{k = 1}^{2\#_{P@W}}$, it is reasonable to directly match such features without using the magnitude-phase hierarchical matching, due to the less complexity by $2\#_{P@W} \ll \#_K$. In terms of NN testing, the proposed I2NN strategy will also be used in the phrase-level matching.

An illustration of phrase-level feature matching is given in Fig. 7. On the left, a matched pair of geometric visual phrases is selected as an example, where yellow indicates the central words and magenta indicates the side words. On the right, the phrase-level feature matching results are shown. Compared with the word-level results in Fig. 4, the number of incorrect pairs is greatly reduced, while the correct ones in the copy-move regions are well preserved. This should be attributed to the enhanced description ability of the phrase-level features.
\begin{figure}[!t]
	\centering
	\subfigure[Matched geometric visual phrases]{\includegraphics[width=1.6in]{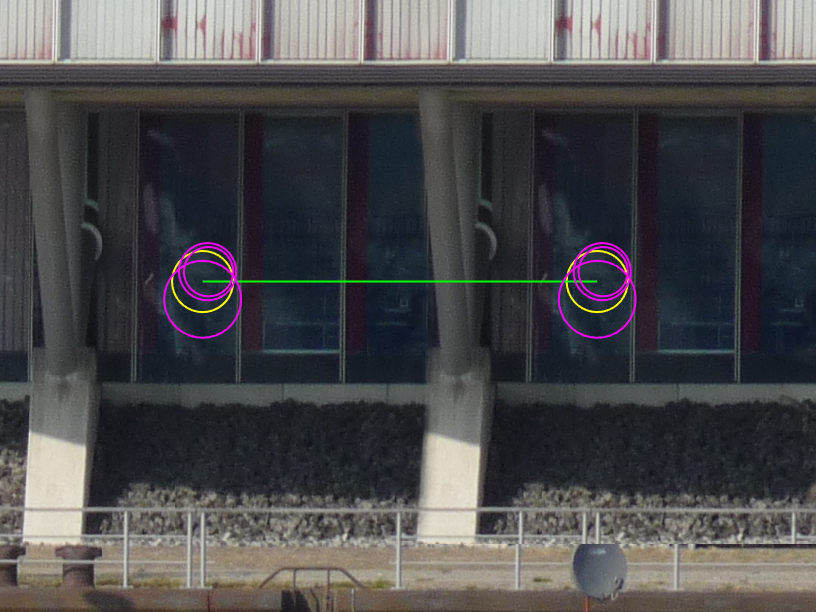}}
	\subfigure[Phrase-level matching results]{\includegraphics[width=1.6in]{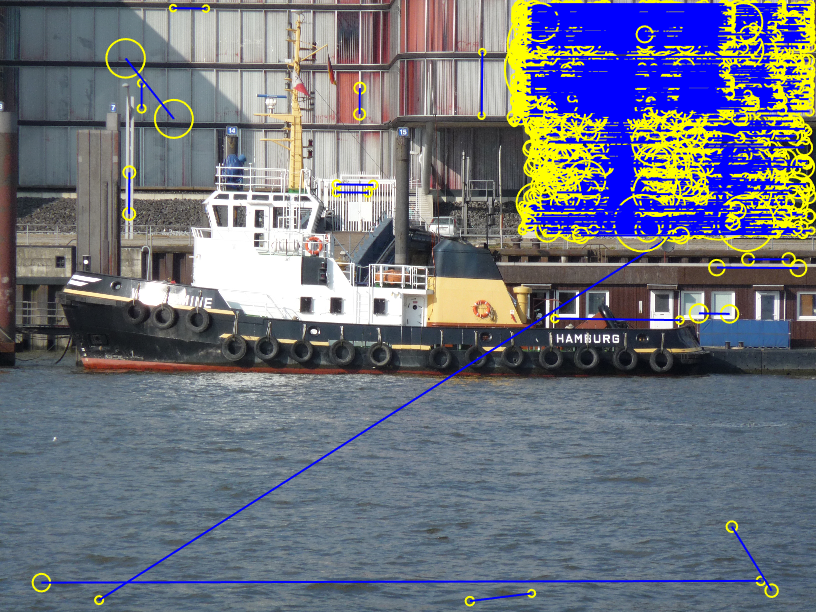}}
	
	\centering
	\caption{Phrase-level feature matching.}
\end{figure}

\subsection{Post-processing}

In this section, the sparse matches will be transformed into the dense localization results through post-processing. Specifically, the proposed post-processing algorithm includes the following steps:
\begin{itemize}
\item \textbf{Step1:} First-stage filtering based on geometric constraints;
\item \textbf{Step2:} Adaptive clustering of matched pairs based on offsets;
\item \textbf{Step3:} Second-stage filtering based on image content;
\item \textbf{Step4:} Forgery localization based on homography and local similarity;
\item \textbf{Step5:} Forgery localization based on matched keypoint regions;
\item \textbf{Step6:} Fusion of the detection results.
\end{itemize}

Here, the detailed description of Steps 1 to 4 can be found in our previous work \cite{ref30}. As a new concept, the combination of Steps 4 to 6 is able to provide better localization recall and precision, which is not available in previous methods. Next, the research motivation and technical details of such new components are presented.

Currently, typical localization strategies in the post-processing stage are mainly divided into the following two categories:
\begin{itemize}
	\item \textbf{Localization strategy based on homography and local similarity} \cite{ref30, ref31}. It first evaluates the geometric transformation relationship of each pair of copy-move regions. Then, the same transformation performed forward and backward on the entire image, allowing the copy-move regions to overlap. Finally, by a pixel-by-pixel correlation check, the regions with stronger correlation are highlighted as the localization results. Ideally, this strategy is able to offer accurate forgery regions with clear boundary. However, due to the smoothness and self-similarity of natural images, high correlation response often appears in the similar-but-genuine regions, leading to false positives.
	\item \textbf{Localization based on matched keypoint regions} \cite{ref19}. It generally takes the union of matched keypoint regions as the basic localization results, and then through certain enhancement technology like morphological processing to improve the accuracy. Obviously, this method has a lower cost because it does not require the dense operations. However, when noise-like attacks exist, the number of matched keypoints may be significantly reduced, leading to false negatives.
\end{itemize}

\begin{figure}[!t]
	\centering
	\subfigure[ROI heat map]{\includegraphics[width=1.1in]{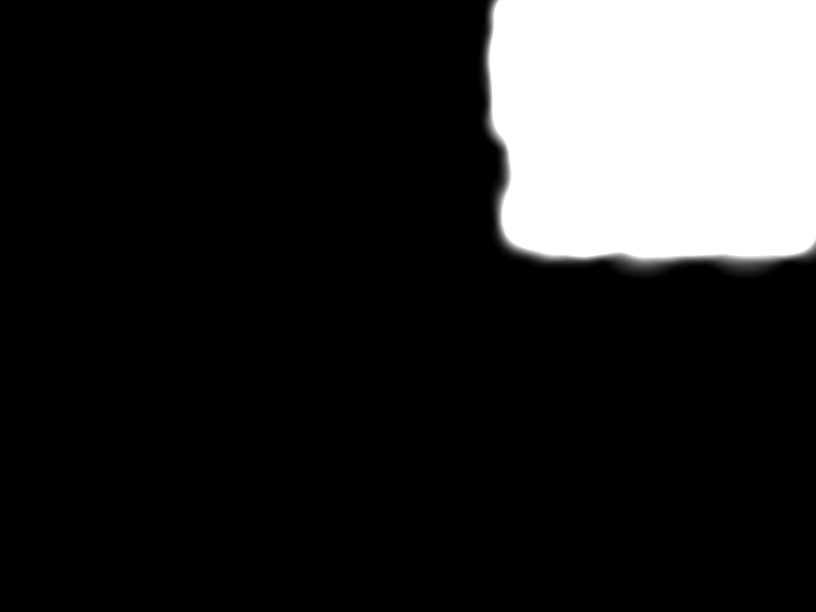}}
	\subfigure[SSIM map]{\includegraphics[width=1.1in]{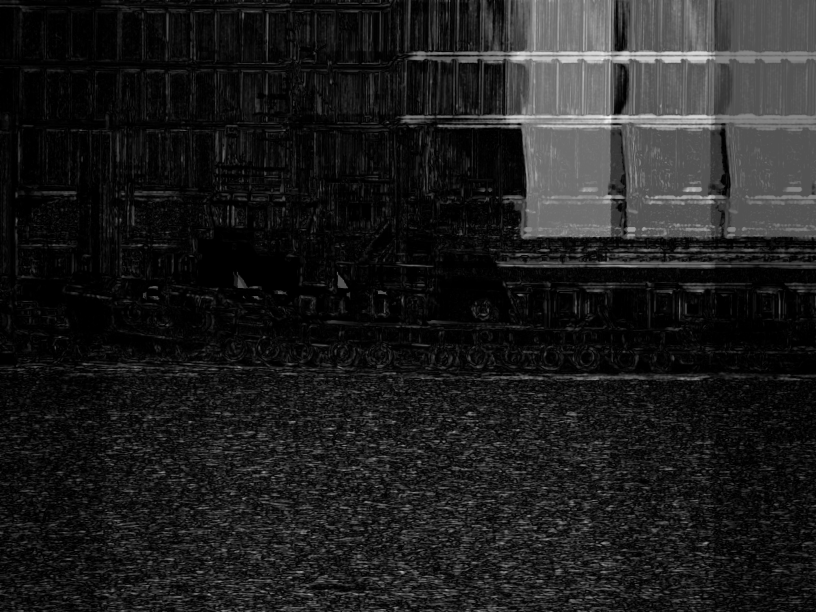}}
	\subfigure[Fusion result]{\includegraphics[width=1.1in]{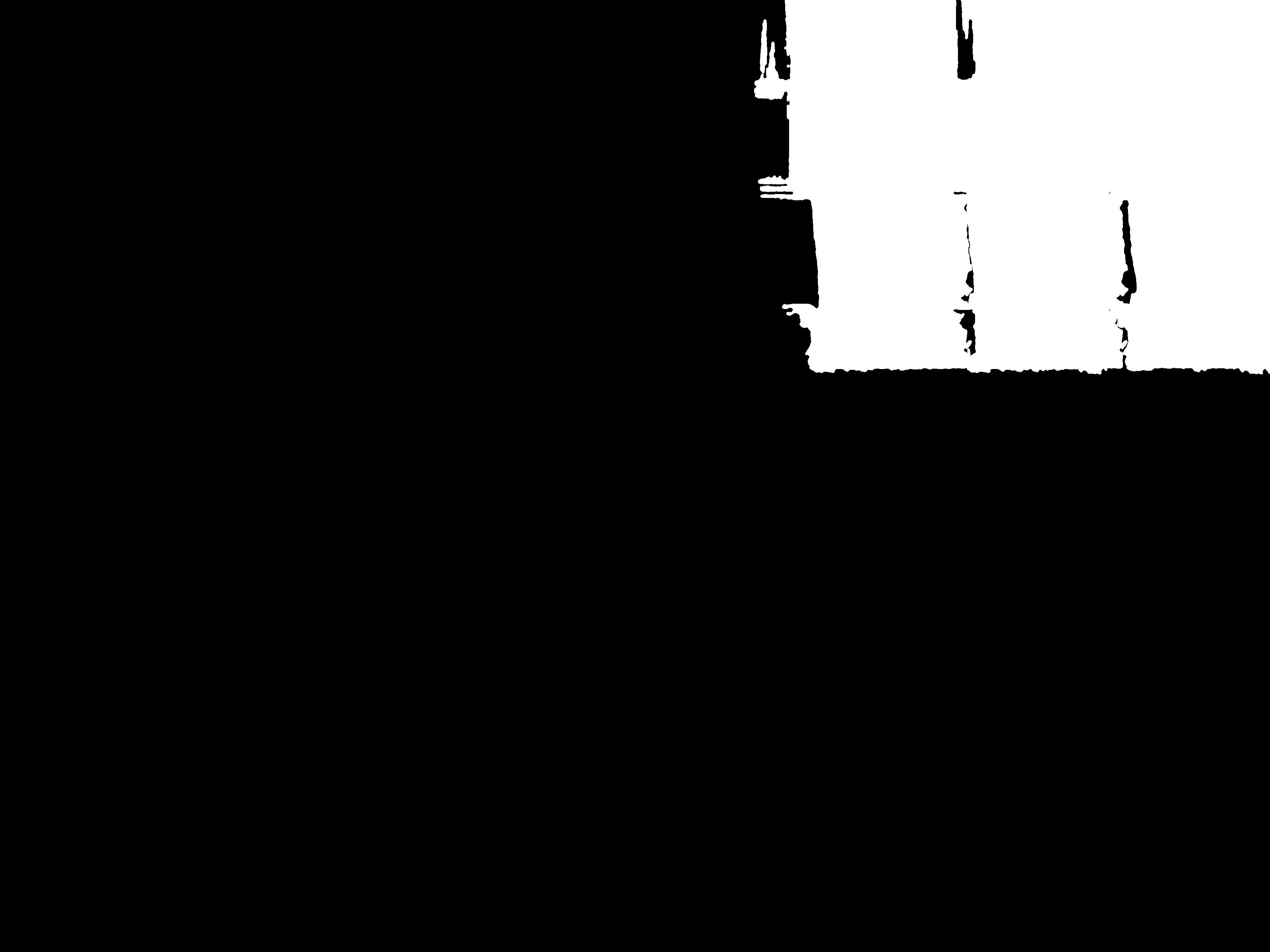}}
	\centering
	\caption{Forgery localization using the fusion strategy.}
\end{figure}

Taken together, the above analysis shows that both methods have difficulties in achieving satisfactory localization accuracy. This is especially true especially when input image is with high self-similarity or strong signal corruption. For this reason, we proposes a \textbf{fusion-based localization} algorithm using the information from homography, local similarity, and keypoint region, with ability to reduce both false positives and false negatives.

To clarify, we denote the set of phrase-level matched pairs as ${{\bf{P}}^p} = \{ {\bf{p}}_k^p = {{\rm{(}}{\bf{k}},{\bf{k'}}{\rm{)}}_k}\} _{k = 1}^{\#_{P@P}}$ where $\#_{P@P}$ is the number of pairs at the phrase level, and the corresponding set of phrase-level matched keypoints as ${{\bf{{K}}}^{p}} = \{ {{\bf{k}}_k}\} _{k = 1}^{2\#_{P@P}} \in {{\bf{{K}}}^{w}}$. For the keypoint ${{\bf{k}}_k} = ({x_k},{y_k},{\sigma _k})$ in the set ${\bf{{K}}}^{p}$, the \textbf{region of interest} (ROI) is defined as $ \{ \gamma_{k} (i,j):(i,j) \in [1,...,{N_I}] \times [1,...,{M_I}]\} $, which satisfies:
\begin{equation}
	\gamma_{k} (i,j) = \left\{ {\begin{array}{*{20}{c}}
			1,&{{{(i - {x_k})}^2} + {{(j - {y_k})}^2} \le {{({\sigma _k} \cdot Z)}^2}}\\
			0,&{{{(i - {x_k})}^2} + {{(j - {y_k})}^2} > {{({\sigma _k} \cdot Z)}^2}}
	\end{array}} \right.,
\end{equation}
where the keypoint ROI map $\gamma_{k}$ has the same resolution ${N_I} \times {M_I}$ as the original image $I$; $\gamma_{k}(i,j) \in \{ 0,1\} $ reflects whether the position $(i,j)$ belongs to the ROI of keypoint ${{\bf{k}}_k}$; $Z$ is the size of the keypoint ROI. By accumulating the keypoint ROI maps, also applying the morphological dilation ${\cal D}$ and Gaussian blurring, the \textbf{ROI heat map} $\{ \Gamma_{k} (i,j):(i,j) \in [1,...,{N_I}] \times [1,...,{M_I}]\} $ for whole image can be obtained:
\begin{equation}
	\begin{split}
		&\Gamma_{k} = {G_{ij}}(i',j') \otimes {\cal \mathcal{D}}\left( {\sum\limits_{k = 1}^{2\#_{P@P}} {\gamma_{k}} } \right)\\
		& \approx \sum\limits_{(i',j') \in S} {{G_{ij}}(i',j') \cdot {\cal \mathcal{D}}\left( {\sum\limits_{k = 1}^{2\#_{P@P}} {\gamma_{k}} } \right)} \\
		& {\rm{ = }}{\mathcal{F}^{ - 1}}(\mathcal{F}({\cal \mathcal{D}}\left( {\sum\limits_{k = 1}^{2\#_{P@P}} {\gamma_{k}} } \right))\mathcal{F}({G_{L,L}}(i',j'))), (i',j') \in S,
	\end{split}
\end{equation}
where the Gaussian kernel ${G_{ij}}(i',j')$ and the enumeration region $S$ are defined in Section \uppercase\expandafter{\romannumeral3}-D2, and the similar fast calculation is also used; the morphological dilation ${\cal D}$ is based on a circular element with radius of ${T_{\cal D}}$ (${T_{\cal D}} = 50$ in our implementation). Subsequently, the response in the ROI heat map is normalized to $[0,1]$, resulting in a \textbf{normalized ROI heat map} $\{ \overline \Gamma (i,j):(i,j) \in [1,...,{N_I}] \times [1,...,{M_I}]\} $:
\begin{equation}
	\overline \Gamma  (i,j) = \left\{ {\begin{array}{*{20}{c}}
			{\frac{{\Gamma (i,j)}}{{{T_{nor}}}}}&{\Gamma (i,j) \le {T_{nor}}}\\
			1&{\Gamma (i,j) > {T_{nor}}}
	\end{array}} \right.,
\end{equation}
where ${T_{nor}}$ is the normalization threshold (${T_{nor}} = 25000$ in our implementation).

To clarify, we denote the matched pair clusters (for managing multiple forgeries, see also \cite{ref30}) as $\{ {{\bf{C}}_k}\} _{k = 1}^{\#_C}$ where $\#_C$ is the number of clusters. The \textbf{SSIM correlation map} (for forgery localization, see also \cite{ref30}) ${\rho _k}$ under cluster ${{\bf{C}}_k}$ is weighted by the normalized ROI heat map $\overline \Gamma $ to merge the information of homography, local similarity, and keypoint region:
\begin{equation}
	{\rho '_k}(i,j) = {\rho _k}(i,j) \cdot \overline \Gamma  (i,j).
\end{equation}

Next, based on the correlation threshold ${T_{cor}}$ (${T_{cor}} = 0.4$ in our implementation), ${\rho '_k}$ is binarized into the localization result ${\tilde \rho '_k}$. After performing the homography estimation and forgery localization for all clusters, the union of such localization maps, $R = \bigcup\nolimits_{k = 1}^{\#_C} {{{\tilde \rho '}_k}} $, is taken as the final copy-move detection result.

\begin{figure*}[!t]
	\centering
	\includegraphics[width=4.5in]{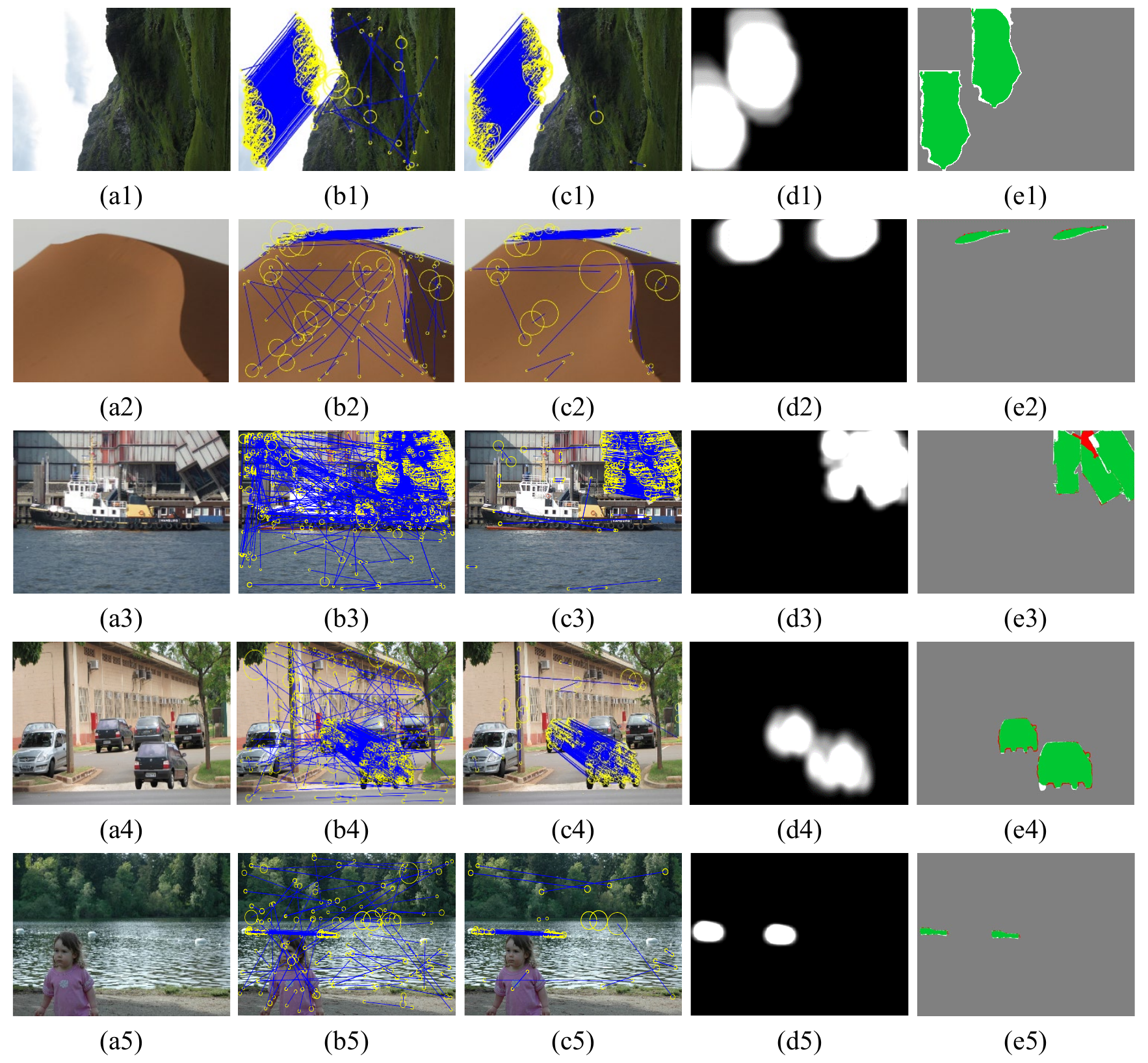}
	\caption{Some examples of the proposed copy-move forgery detection algorithm on FAU, GRIP, MICC, CMH, and UNIPA (from top to bottom). Here, (a1)–(a5) are the tampered images, (b1)–(b5) are the word-level feature matching results, (c1)–(c5) are the phrase-level feature matching results, (d1)–(d5) are the normalized ROI heat maps, and (e1)–(e5) are the final detection results. For (e1)–(e5), the true positives, false positives, and false negatives are marked in green, red, and white, respectively. The results illustrate the higher discriminability of phrase-level feature description and matching w.r.t. word-level ones.}
\end{figure*}

An example of our localization strategy is given in Fig. 8. As can be seen, the normalized ROI heat map is able to draw the potential regions, but such a result is rough. The SSIM map gives a more accurate boundary of the potential regions, but the response values is quite high for some similar-but-genuine regions, tending to raise false positives. For above facts, the proposed strategy merges such two types of results with complementary information, and hence exhibits lower false positives and false negatives.

\section{Performance Evaluation}

In this section, we will conduct a number of benchmark and robustness experiments for evaluating the performance of our work w.r.t. the state of the art. In addition, we also present the performance analysis for key components and under challenging scenarios. Our code is available at \texttt{github.com/ChaoWang1016/word2phraseCMFD}.

\subsection{Experimental Settings}

The following experiments mainly involve seven copy-move forgery detection datasets: FAU \cite{ref2}, GRIP \cite{ref17}, MICC \cite{ref35}, CMH \cite{ref34}, UNIPA \cite{ref40}, CASIA \cite{ref43}, and CoMoFoD \cite{ref44}. 

The first five datasets are smaller in scale, with the number of images ranging from 48 to 159. Regarding the operations, FAU, GRIP, and UNIPA contain only rigid copy-move manipulation; while MICC and CMH are with further attacks (mainly scaling and rotation) for a convincing visual effect. The last two datasets are larger in scale, with 1313 and 5000 images for CASIA  and CoMoFoD, respectively, while covering more challenging scenarios, e.g., multiple forgeries, strong geometric/signal distortions, and similar-but-genuine backgrounds.

The F1 score is introduced as a main quantitative indicator of detection accuracy. The Precision and Recall scores are also considered as additional accuracy indicators, reflecting false positives and false negatives, respectively. Unlike previous studies, the ground-truth mask will be binarized when calculating such scores, leading to a more rigorous performance evaluation.

\begin{table}
	\centering
	\caption{Precision, Recall, and F1 scores (\%) for Main Competitors on Multiple Datasets.}
	\begin{tabular}{ccccccc}
		\toprule
		\multicolumn{2}{c}{Method} & \multicolumn{1}{c}{\tabincell{c}{PM \\ \cite{ref17}}} & \multicolumn{1}{c}{\tabincell{c}{Iteration \\ \cite{ref27}}} & \multicolumn{1}{c}{\tabincell{c}{HFPM \\ \cite{ref19}}} &
		\multicolumn{1}{c}{\tabincell{c}{CMI \\ \cite{ref30}}} & \multicolumn{1}{c}{Proposed} \\
		\midrule
		\multicolumn{1}{c}{\multirow{6}[1]{*}{\begin{sideways}Precision\end{sideways}}} 
		& FAU & 93.1  & 79.6  & 89.4  & 87.0  & 93.5   \\ 
		& GRIP & 93.0  & 81.9  & 92.7  & 87.7  & 92.3   \\ 
		& MICC & 92.0  & 66.1  & 88.8  & 86.5 & 92.5   \\ 
		& CMH & 83.0  & 55.0  & 85.3  & 79.8  & 84.4   \\ 
		& UNIPA & 85.4  & 73.1  & 86.2  & 89.4  & 93.9   \\
		& \textbf{Avg.} & \textbf{89.3} & \textbf{71.1} & \textbf{88.5} & \textbf{86.1} & \textbf{91.3} \\
		\midrule
		\multicolumn{1}{c}{\multirow{6}[1]{*}{\begin{sideways}Recall\end{sideways}}} 
		& FAU & 91.9  & 95.1  & 88.9  & 91.3  & 92.8   \\ 
		& GRIP & 96.4  & 98.6  & 97.4  & 99.7  & 98.5   \\ 
		& MICC & 89.1  & 75.0  & 86.5  & 90.2 & 89.2   \\ 
		& CMH & 79.0  & 65.3  & 72.0  & 88.5  & 81.4   \\ 
		& UNIPA & 91.8  & 99.1  & 94.0  & 99.1  & 95.9   \\
		& \textbf{Avg.} & \textbf{89.6} & \textbf{86.6} & \textbf{87.8} & \textbf{93.8} & \textbf{91.6} \\
		\midrule
		\multicolumn{1}{c}{\multirow{6}[1]{*}{\begin{sideways}F1\end{sideways}}} 
		& FAU & 91.9  & 83.4  & 88.6  & 88.0  & 91.3   \\ 
		& GRIP & 93.9  & 86.7  & 94.8  & 92.5  & 93.9   \\ 
		& MICC & 89.2  & 66.4  & 86.7  & 87.9 & 88.2   \\ 
		& CMH & 80.1  & 58.3  & 76.4  & 80.3  & 82.2   \\ 
		& UNIPA & 88.3  & 81.0  & 89.3  & 89.8  & 94.4   \\
		& \textbf{Avg.} & \textbf{88.7} & \textbf{75.2} & \textbf{87.2} & \textbf{87.7} & \textbf{90.0} \\
		\bottomrule
	\end{tabular}%
	\label{tab:addlabel}%
\end{table}%

\begin{table*}[bp]
	\caption{Precision, Recall, and F1 scores (\%) for Comprehensive Competitors on FAU Dataset.}
	\centering
	\begin{threeparttable}
		\begin{tabular}{cccccccccccc}
			\toprule
			Metrics &\tabincell{c}{PM \\ \cite{ref17}} & \tabincell{c}{FROGS \\ \cite{ref41}} & \tabincell{c}{ECSS \\ \cite{ref13}} & \tabincell{c}{Segment \\ \cite{ref42}} &\tabincell{c}{Iteration \\ \cite{ref27}} &\tabincell{c}{HFPM \\ \cite{ref19}} & \tabincell{c}{CMI \\ \cite{ref30}} &\tabincell{c}{DeepNet \\ \cite{ref45}} &\tabincell{c}{BusterNet \\ \cite{ref46}} &\tabincell{c}{DenseNet \\ \cite{ref4}} &Proposed \\
			\midrule
			Precision & 93.1 & 90.1 & 90.6 & 88.0 & 79.6 & 89.4 & 87.0 & 30.3 & 44.5 & 75.4 & 93.5 \\
			Recall & 91.9 & 90.4 & 91.5 & 80.9 & 95.1 & 88.9 & 91.3 & 26.3 & 31.5 & 73.9 & 92.8 \\
			F1 & 91.9 & 90.2 & 90.9 & 84.4 & 83.4 & 88.6 & 88.0 & 28.1 & 36.9 & 74.7 & 91.3 \\ 
			\bottomrule
		\end{tabular}
	\end{threeparttable}
\end{table*}

\subsection{Benchmark Experiments}
In this round of experiments, we provide the performance statistics on five benchmarks for positioning our algorithm w.r.t. the current state-of-the-art approaches.

\emph{Qualitative Cases.} In Fig. 9, some detection samples on FAU, GRIP, MICC, CMH, and UNIPA are illustrated. It can be seen that the number of false matches in (c1)–(c5) is significantly reduced compared to (b1)–(b5), while the correct matches in the copy-move regions are well preserved. This phenomenon indicates that the proposed phrase-level features can better describe the image local behavior, compared to the word-level features used in traditional copy-move detection methods. In addition, the normalized ROI heat maps in (d1)–(d5) are roughly consistent with the ground-truth forgery regions; the fusion localization strategy is able to give more accurate results (e1)–(e5), i.e., with high precision and high recall at the same time.

\begin{table*}[h]
	\centering
	\caption{Precision, Recall, and F1 scores (\%) for Different Settings on FAU Dataset}
	\begin{tabular}{ccccccccc}
		\toprule
		\multicolumn{2}{c}{\tabincell{c}{Parameter Setting}} & \tabincell{c}{\#1 } & \tabincell{c}{\#2 } & \tabincell{c}{\#3 } &
		\tabincell{c}{\#4 } & \tabincell{c}{\#5} & \tabincell{c}{\#6}& \tabincell{c}{\#7}\\
		\midrule
		\multicolumn{1}{c}{\multirow{3}[1]{*}{$K$}} & 3 & \checkmark & ~ & ~ & \checkmark & \checkmark & \checkmark & \checkmark \\
		& 1 & ~ & \checkmark & ~ & ~ & ~ & ~ & ~ \\
		& 5 & ~ & ~ & \checkmark & ~ & ~ & ~ & ~ \\
		\midrule
		\multicolumn{1}{c}{\multirow{3}[1]{*}{$T_{sml{\rm{-}}rat}$}} & 0.6 & \checkmark & \checkmark & \checkmark & ~ & ~ &  \checkmark & \checkmark \\
		& 0.5 & ~ & ~ & ~ & \checkmark & ~ & ~ & ~ \\
		& 0.7 & ~ & ~ & ~ & ~ & \checkmark & ~ & ~ \\
		\midrule
		\multicolumn{1}{c}{\multirow{3}[1]{*}{${T_{cor}}$}} & 0.4 & \checkmark & \checkmark & \checkmark & \checkmark & \checkmark & ~ & ~ \\
		& 0.3 & ~ & ~ & ~ & ~ & ~ & \checkmark & ~ \\
		& 0.5 & ~ & ~ & ~ & ~ & ~ & ~ & \checkmark \\
		\midrule \midrule
		\multicolumn{1}{c}{\multirow{3}[1]{*}{Scores}} & Precision & 93.5 & 92.4  & 93.2  & 93.7  & 92.9  & 92.1  & 94.2  \\
		& Recall & 92.8 & 93.5  & 92.1  & 92.7  & 93.4  & 93.6  & 92.7 \\
		& F1 & 91.3 & 91.0  & 91.1  & 91.5  & 91.2  & 90.9  & 91.7 \\
		\bottomrule
	\end{tabular}%
\end{table*}%

\emph{Main Competitors.} In Table \uppercase\expandafter{\romannumeral1}, we list the Precision, Recall, and F1 scores on FAU, GRIP, MICC, CMH, and UNIPA. The three most representative copy-move forgery detection algorithms are selected as the reference techniques: PM \cite{ref17}, Iteration \cite{ref27}, and HFPM \cite{ref19}. Here, the previous word-level version for the work in this paper, i.e., CMI \cite{ref30}, is also introduced. The quantitative results in Table \uppercase\expandafter{\romannumeral1} are derived from the code they released. Here, Iteration, HFPM, and CMI are the state-of-the-art algorithms in sparse-field approach; PM is the state-of-the-art algorithm in dense-field approach.

As we mentioned in Section \uppercase\expandafter{\romannumeral2}, the main advantage of the dense-field approach lies in the localization accuracy (especially for rigid translation), while the sparse-field approach provides higher robustness (especially for rotation and scaling). The results in Table \uppercase\expandafter{\romannumeral1} generally support above conclusion, i.e., the PM exhibits relatively better accuracy than Iteration and HFPM in plain copy-move scenarios. Although the proposed method and our previous CMI work in sparse domain, they still provide comparable performance with PM in the Precision, Recall and F1 scores. Meanwhile, our works are generally more accurate compared to the sparse Iteration and HFPM, especially on CMH and UNIPA. Over a variety of benchmarks, the average scores indicate that the proposed method offers the best overall localization accuracy at the levels of Precision, Recall, and F1. The main reasons are as follows. The proposed algorithm is able to deal with small/smooth forgery regions (in GRIP and UNIPA), multiple copy-move forgeries (in FAU and MICC), and rotation/scaling attacks (in MICC and CMH), while better overcoming the semantic gap problem such as high false positives in similar-but-genuine regions w.r.t. word-level CMI. Note that the work in this paper does not show a significant score gap w.r.t. CMI on these small-scale benchmarks, mainly under rigid copy-move manipulation. As illustrated later, such gap will be greater on large-scale benchmarks under challenging scenarios.

\emph{Comprehensive Competitors.} In Table \uppercase\expandafter{\romannumeral2}, we attempt to extend the above benchmark study by introducing more state-of-the-art algorithms, where some of them are closed-source. These newly introduced methods cover: 1) recent developments of dense approach – FROGS \cite{ref41} and ECSS \cite{ref13}, 2) fusion of dense and sparse approaches – Segment \cite{ref42}, and 3) deep learning based approach – DeepNet \cite{ref45}, BusterNet \cite{ref46}, and DenseNet \cite{ref4}. Here, the results on FAU for such methods are cited directly from the reference \cite{ref4}. As expected, the dense methods, FROGS and ECSS, work well under such rigid copy-move scenario. While deep learning based methods, DeepNet, BusterNet, and DenseNet, currently cannot provide comparable scores on FAU w.r.t. hand-crafted dense or sparse methods. Probably due to the lack of prior knowledge about copy-move forgery in their learning process. In general, our method still maintains the advantage w.r.t. these newly introduced works, hence further confirming the effectiveness.

\emph{Parameter Sensitivity.} In general, the proposed algorithm involves numerous parameters, most of which have a clear physical meaning and can be set accordingly. The main parameters and their values in the implementation have been mentioned in the Section \uppercase\expandafter{\romannumeral3}; the rest of the parameter settings can be found in our code (not listed here due to space constraints). In Table \uppercase\expandafter{\romannumeral3}, we summarize the scores for multiple settings to evaluate the parameter sensitivity. More specifically, three main parameters, $K$, $T_{sml{\rm{-}}rat}$, and ${T_{cor}}$, in feature description, matching, and post-processing, are selected. For each parameter, three reasonable values are set, resulting in a total of seven possible combinations. Under such settings, it can be seen that the average scores on FAU fluctuate within a small interval, which is generally acceptable in practice. This phenomenon suggests that our algorithm is not sensitive to certain changes w.r.t. reasonable settings, and the relevant experimental results under setting \#1 in this paper are convincing (i.e., not an isolated case).

\subsection{Robustness Experiments}
In this round of experiments, we provide the performance statistics in the scenarios with geometric transformations or signal corruptions. Specifically, the copy-move images in the benchmark FAU are further attacked by the following four operations: scaling, rotation, additive Gaussian white noise, and JPEG compression. Their parameter settings are given by Table \uppercase\expandafter{\romannumeral4}.

In Fig. 10, we show the Precision, Recall, and F1 curves of various competing algorithms w.r.t. different attack parameters. This part covers all the compared methods in the previous benchmark experiments. Similarly, the scores for closed-source methods are cited directly from the reference \cite{ref4}.

\begin{table}
	\caption{Parameter Settings of the Attacks.}
	\centering
	\begin{tabular}{ccc}
		\toprule
		Attack & Parameter & Range \\
		\midrule
		Scaling & Scale Factor \% & 80, 91 : 2 : 109, 120 \\ 
		Rotation & Degree & 2 : 2 : 10, 20, 60, 180 \\ 
		White Gaussian noise & Standard Deviation & 0.02 : 0.02 : 0.1 \\ 
		JPEG Compression & Quality Factor & 100 : -10 : 20 \\
		\bottomrule
	\end{tabular}
\end{table}

\begin{figure*}[!t]
	\centering
	
	\subfigure{\includegraphics[width=1.4in]{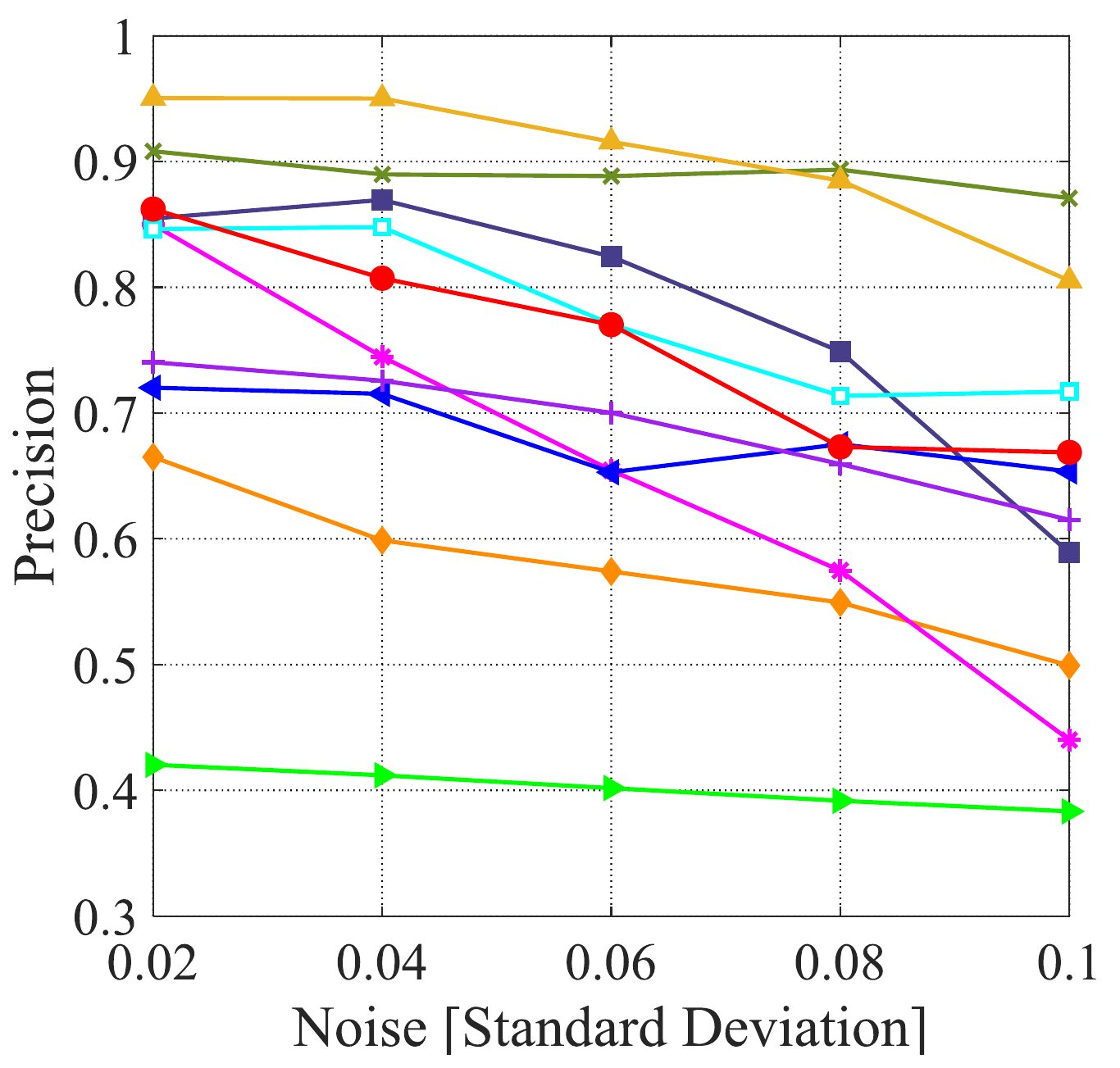}}
	\subfigure{\includegraphics[width=1.4in]{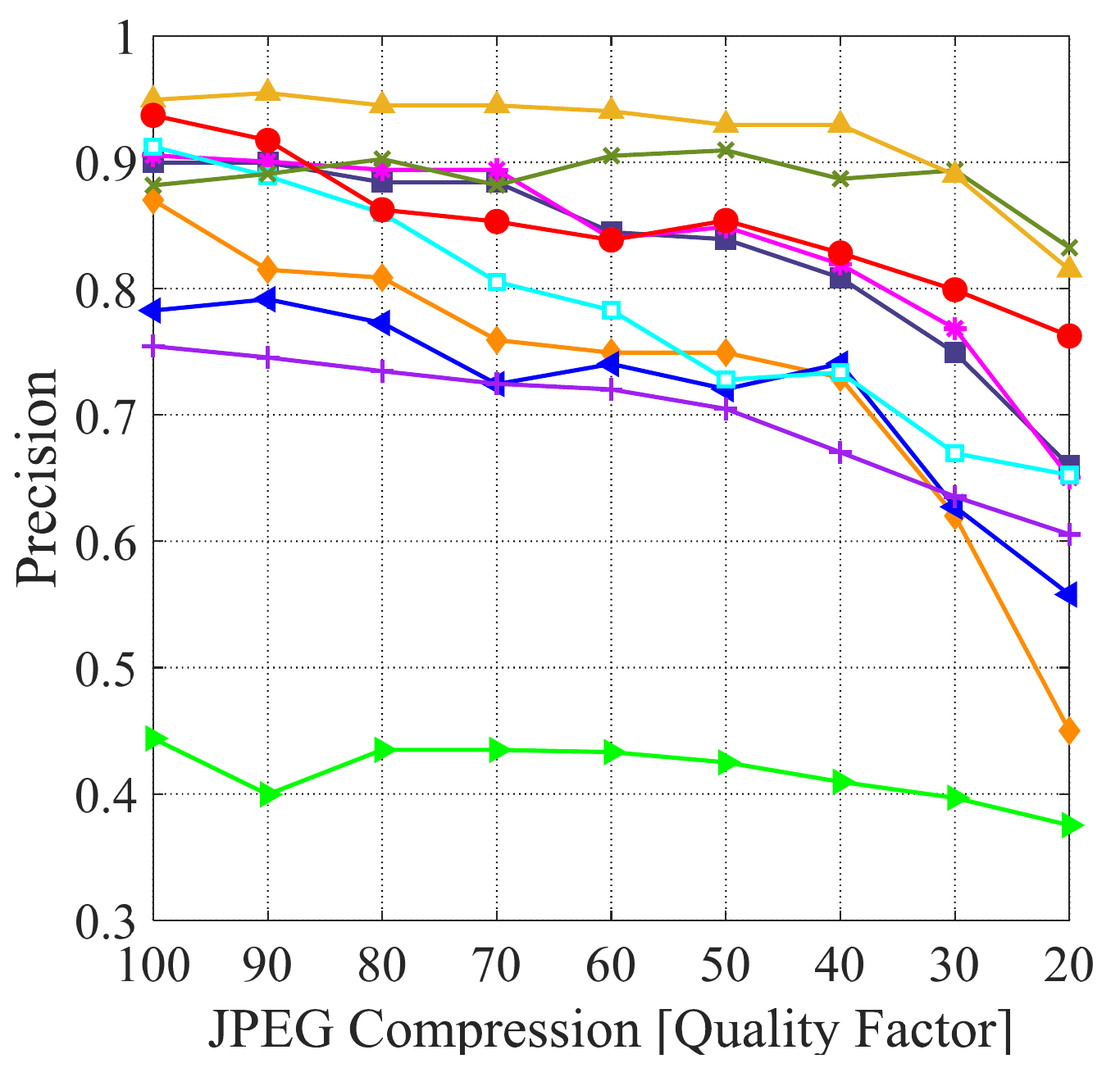}}
	\subfigure{\includegraphics[width=1.4in]{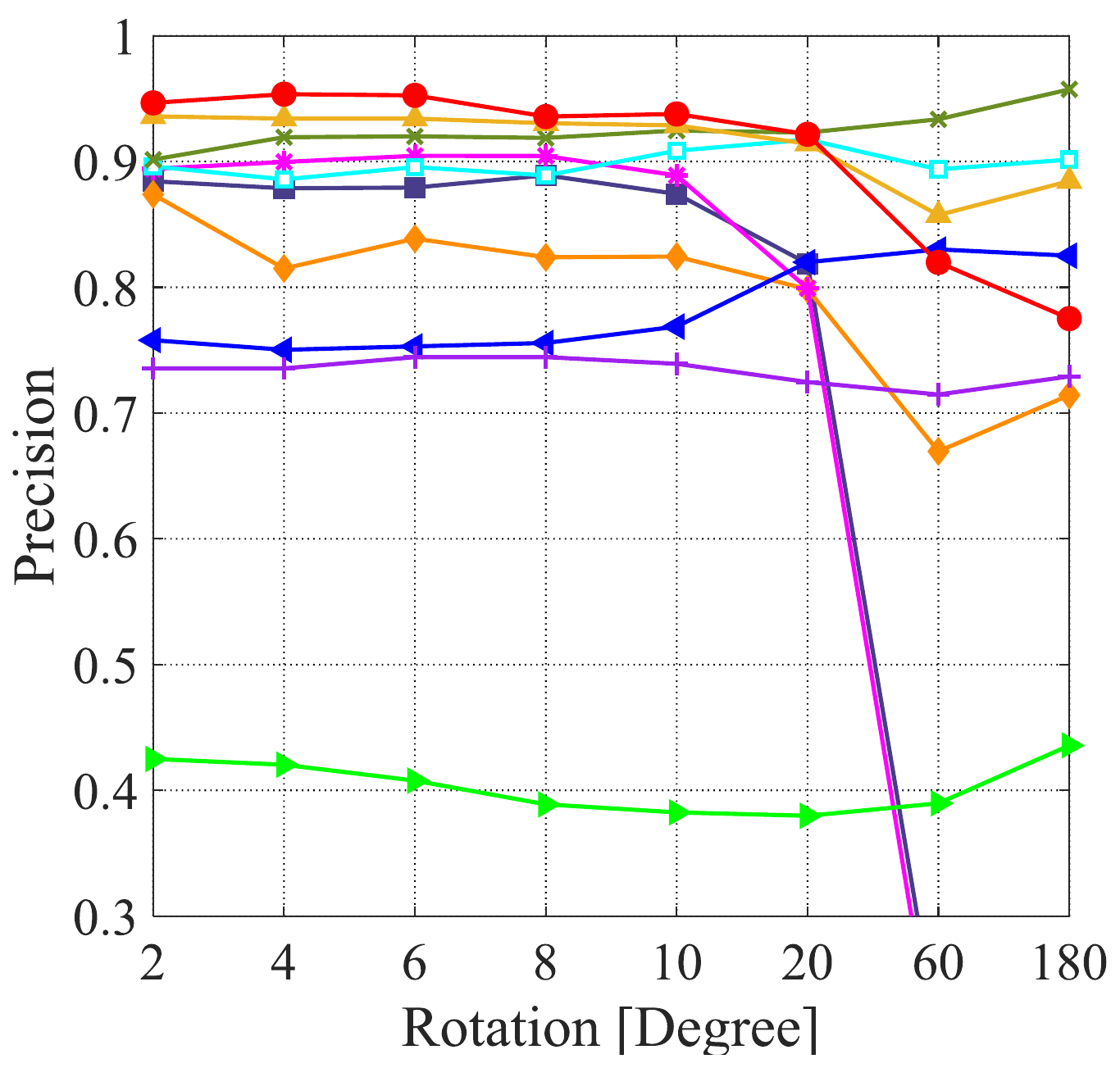}}
	\subfigure{\includegraphics[width=1.4in]{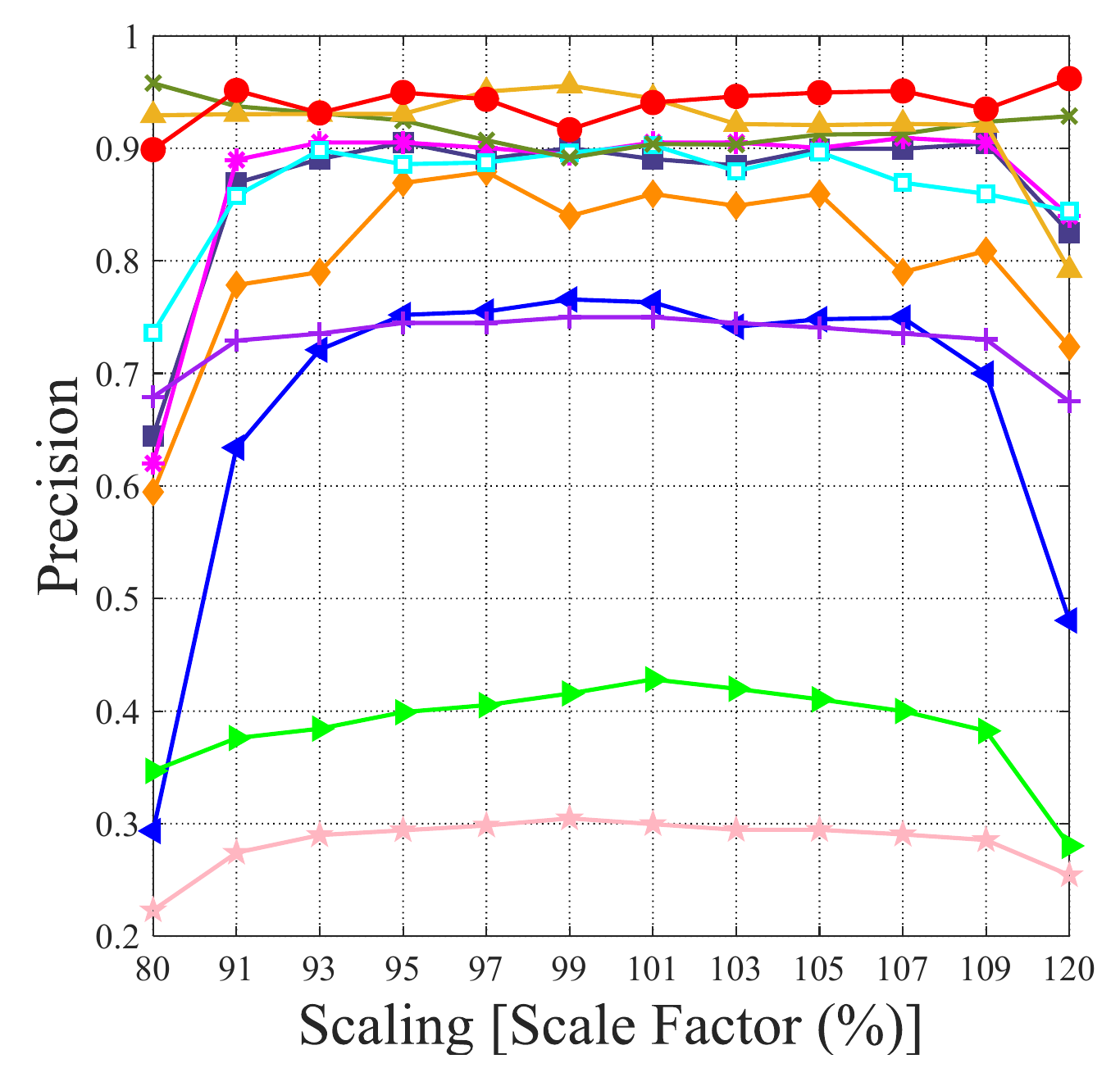}}
	\\
	\subfigure{\includegraphics[width=1.4in]{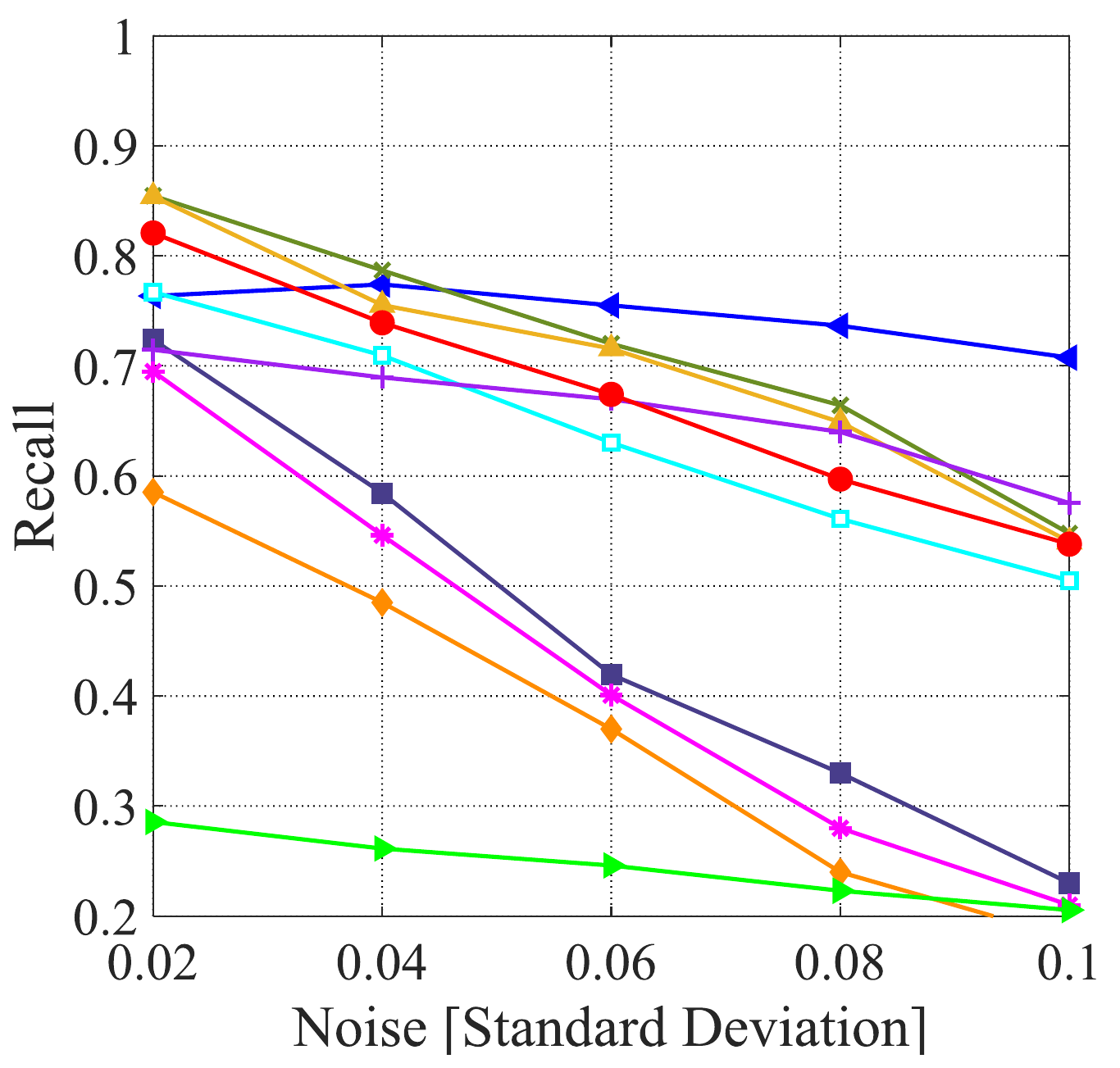}}
	\subfigure{\includegraphics[width=1.4in]{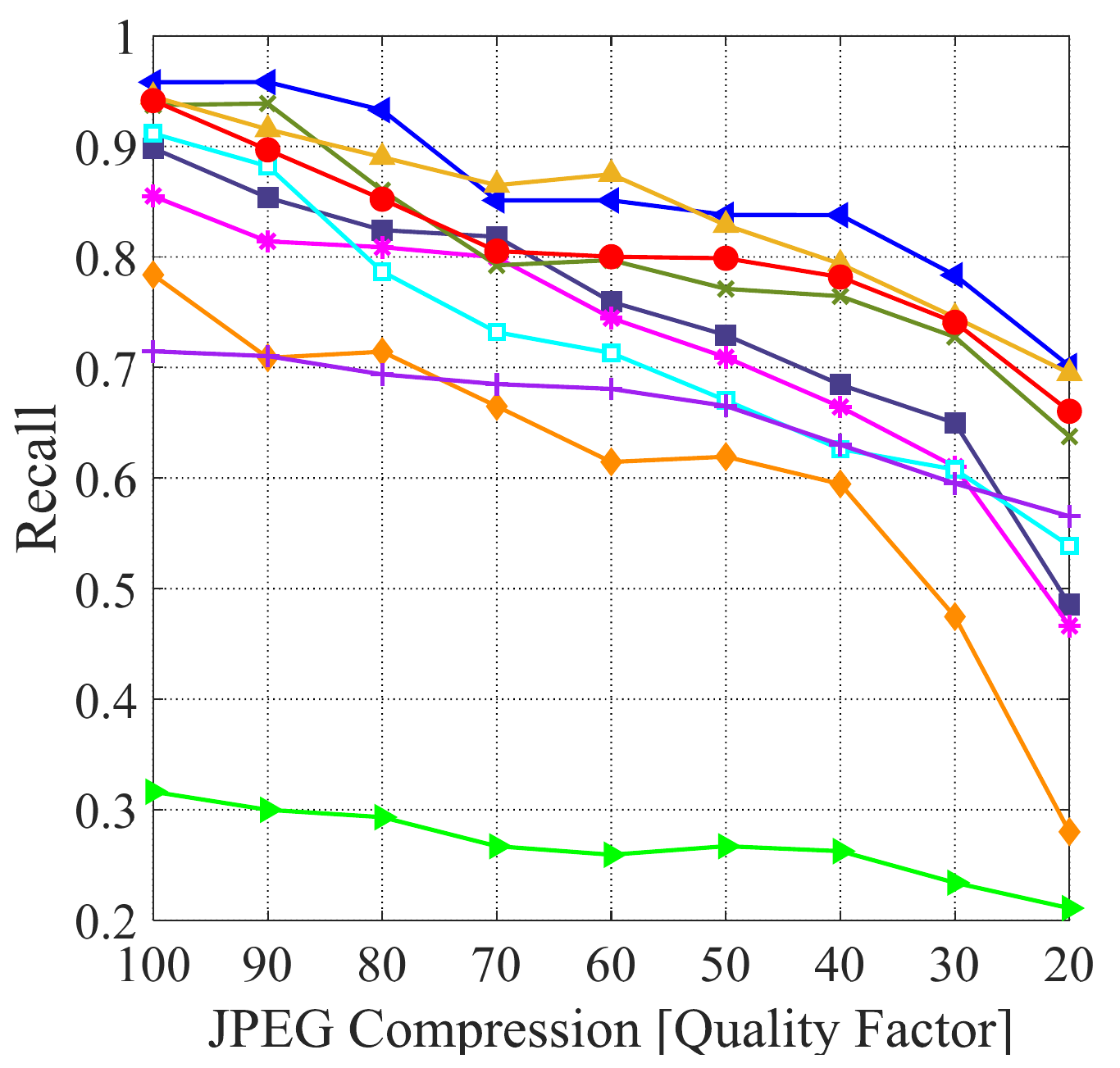}}
	\subfigure{\includegraphics[width=1.4in]{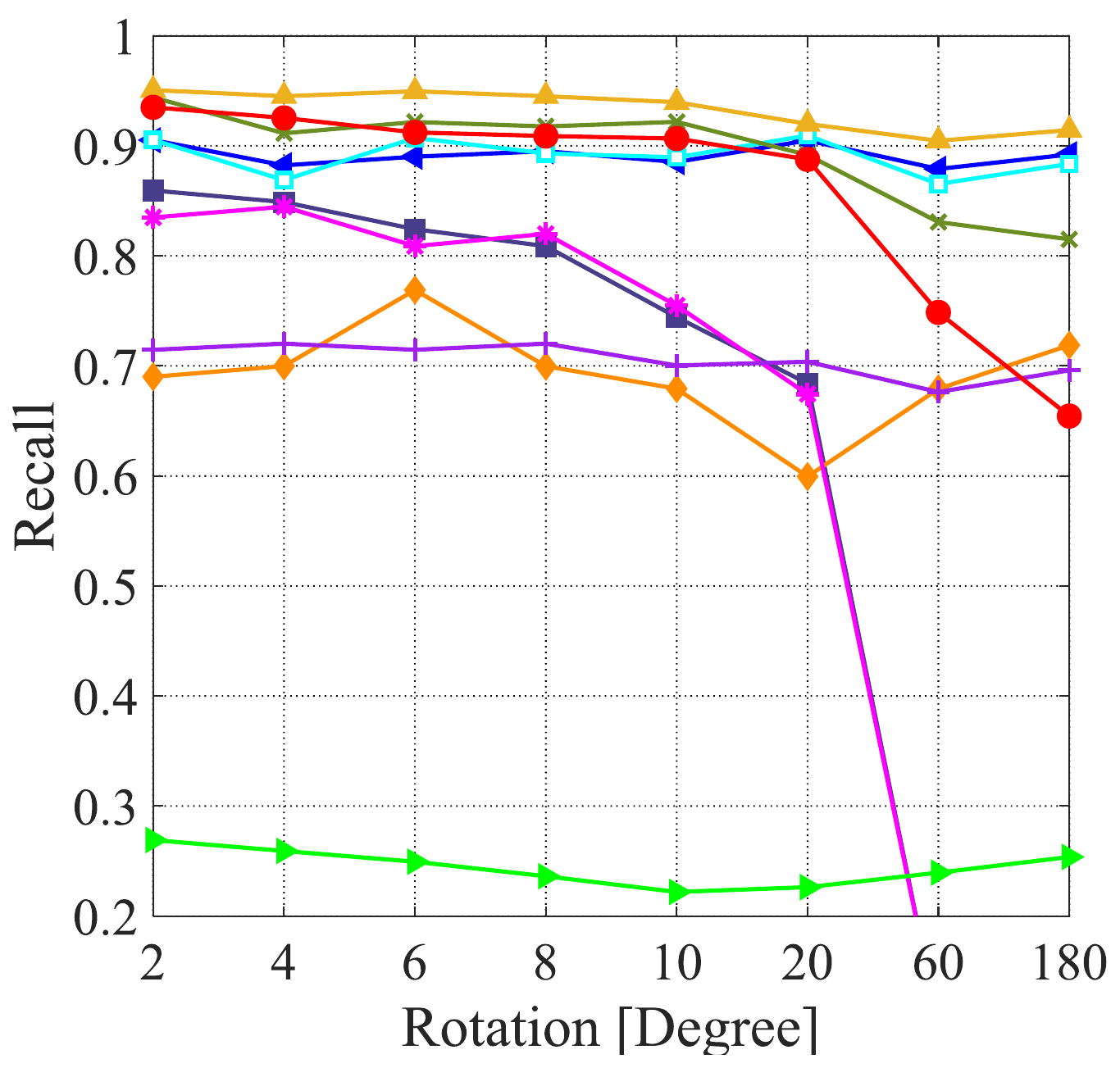}}
	\subfigure{\includegraphics[width=1.4in]{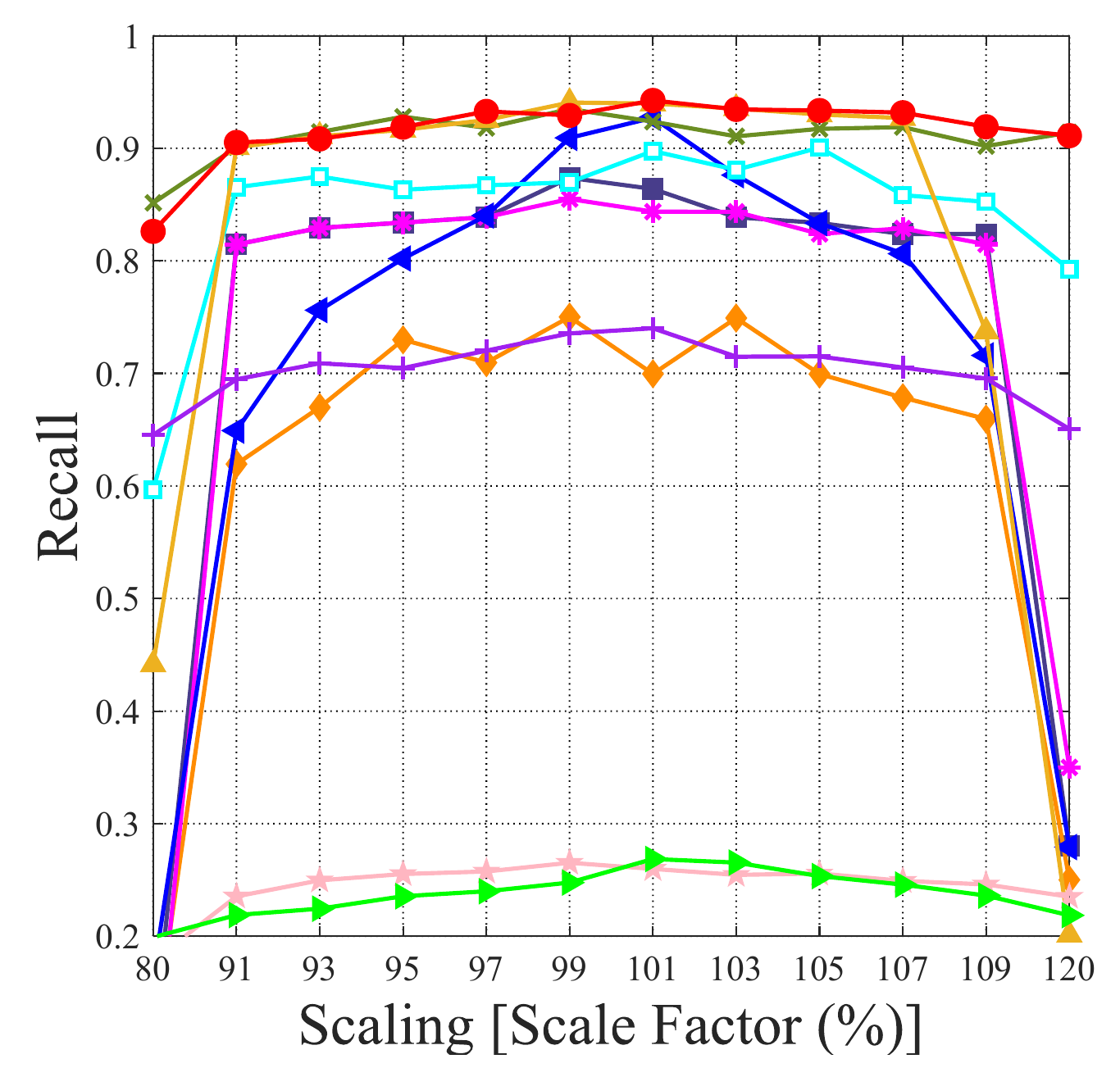}}
	\\
	\subfigure{\includegraphics[width=1.4in]{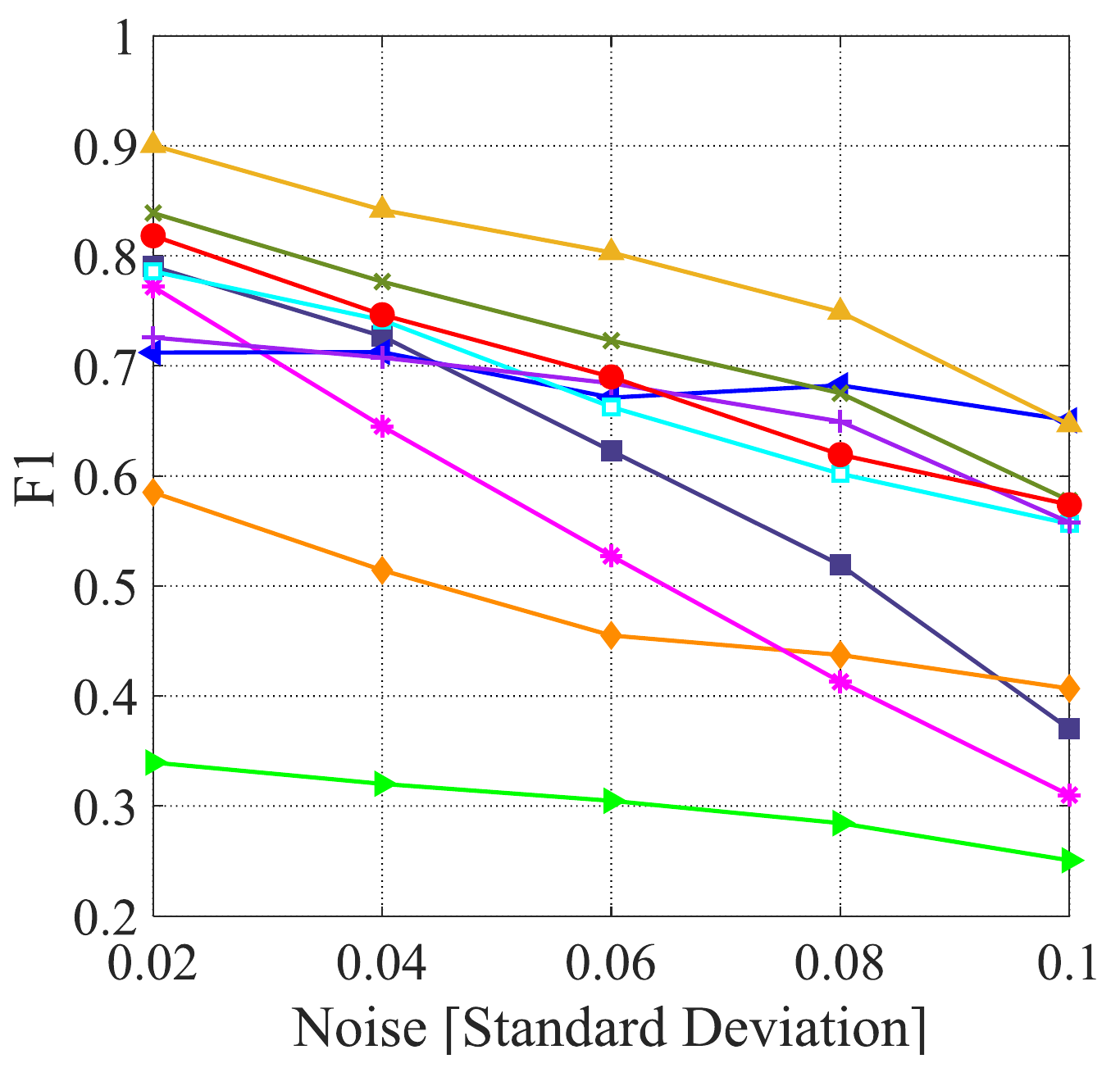}}
	\subfigure{\includegraphics[width=1.4in]{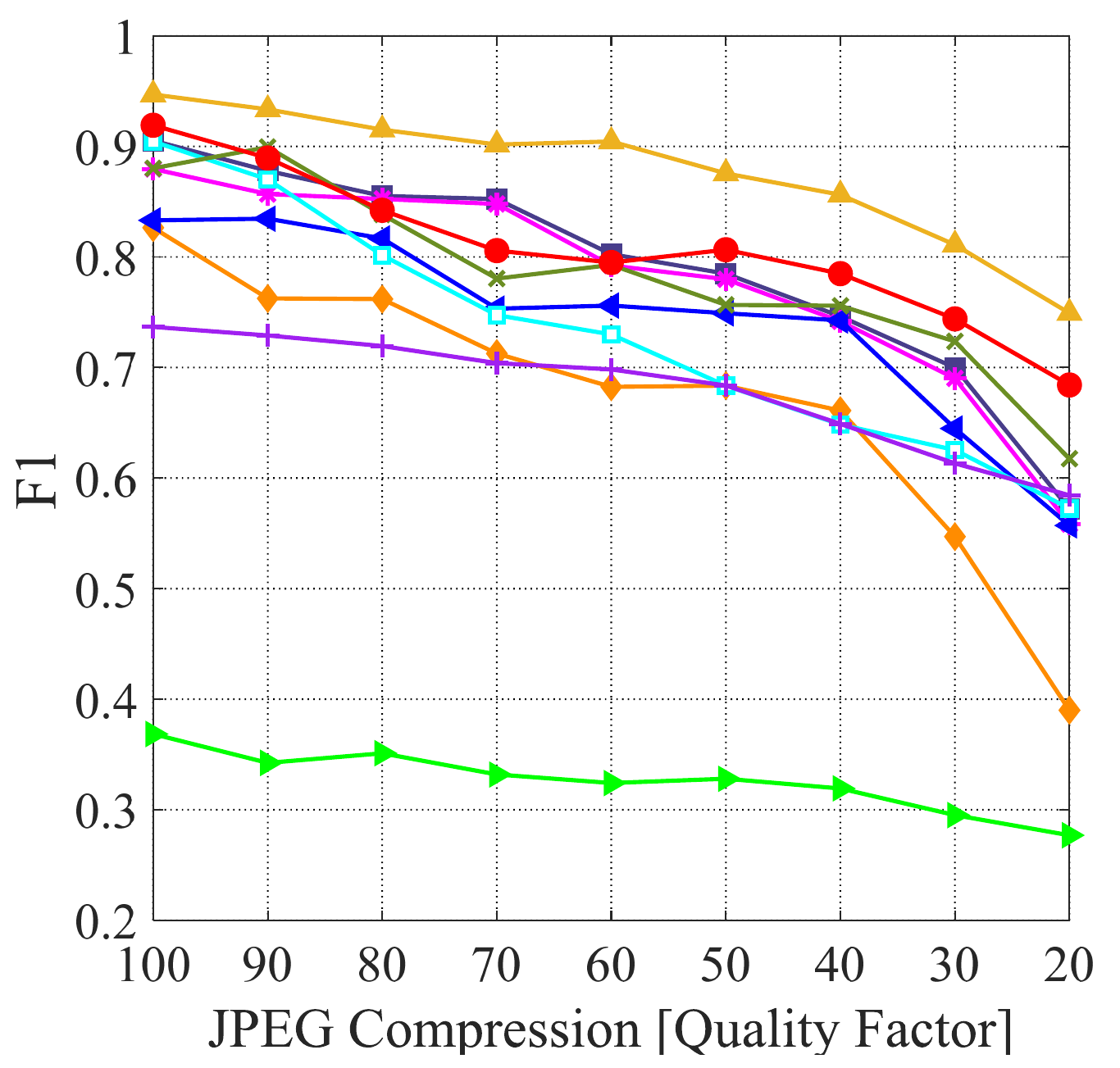}}
	\subfigure{\includegraphics[width=1.4in]{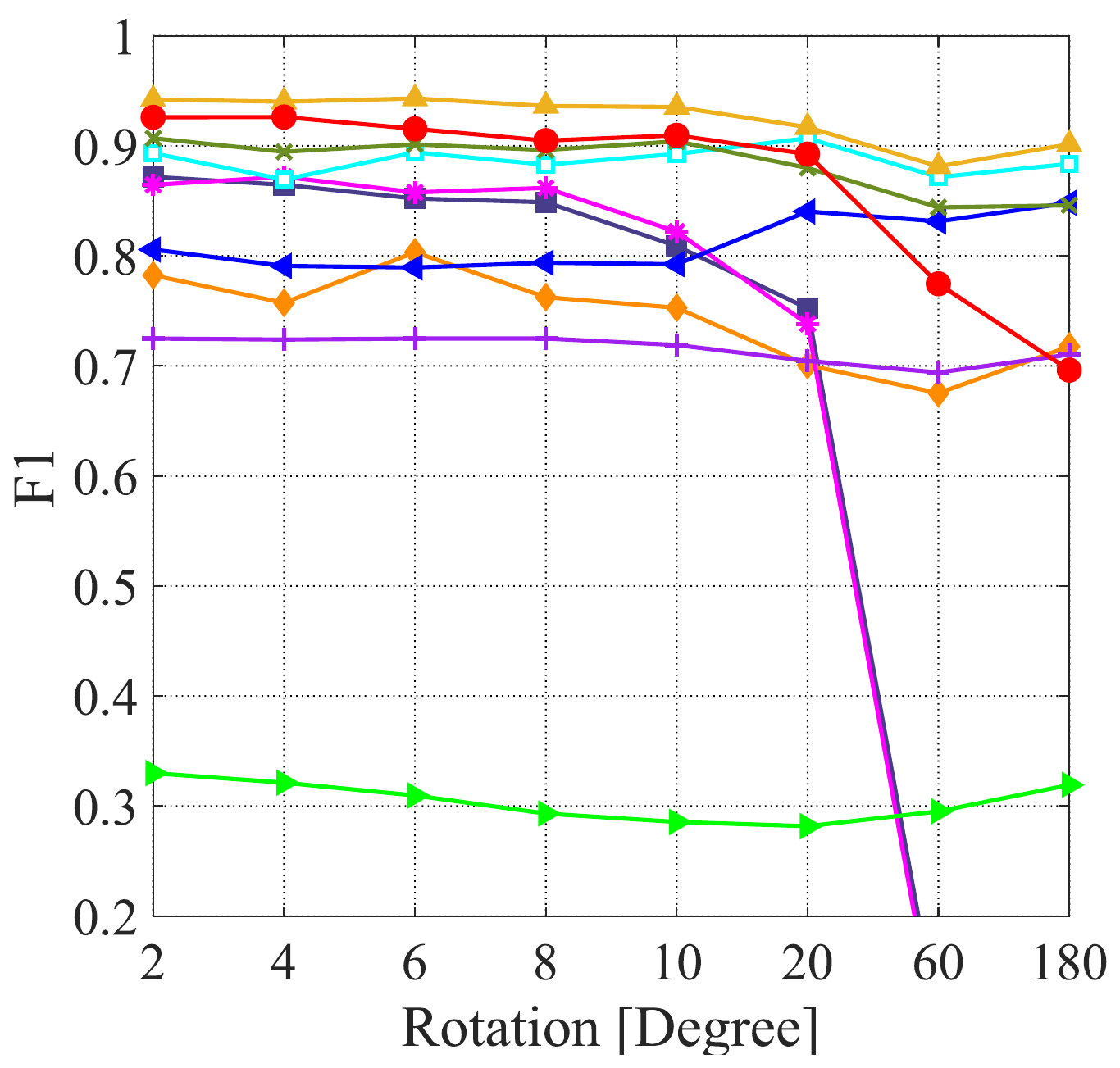}}
	\subfigure{\includegraphics[width=1.4in]{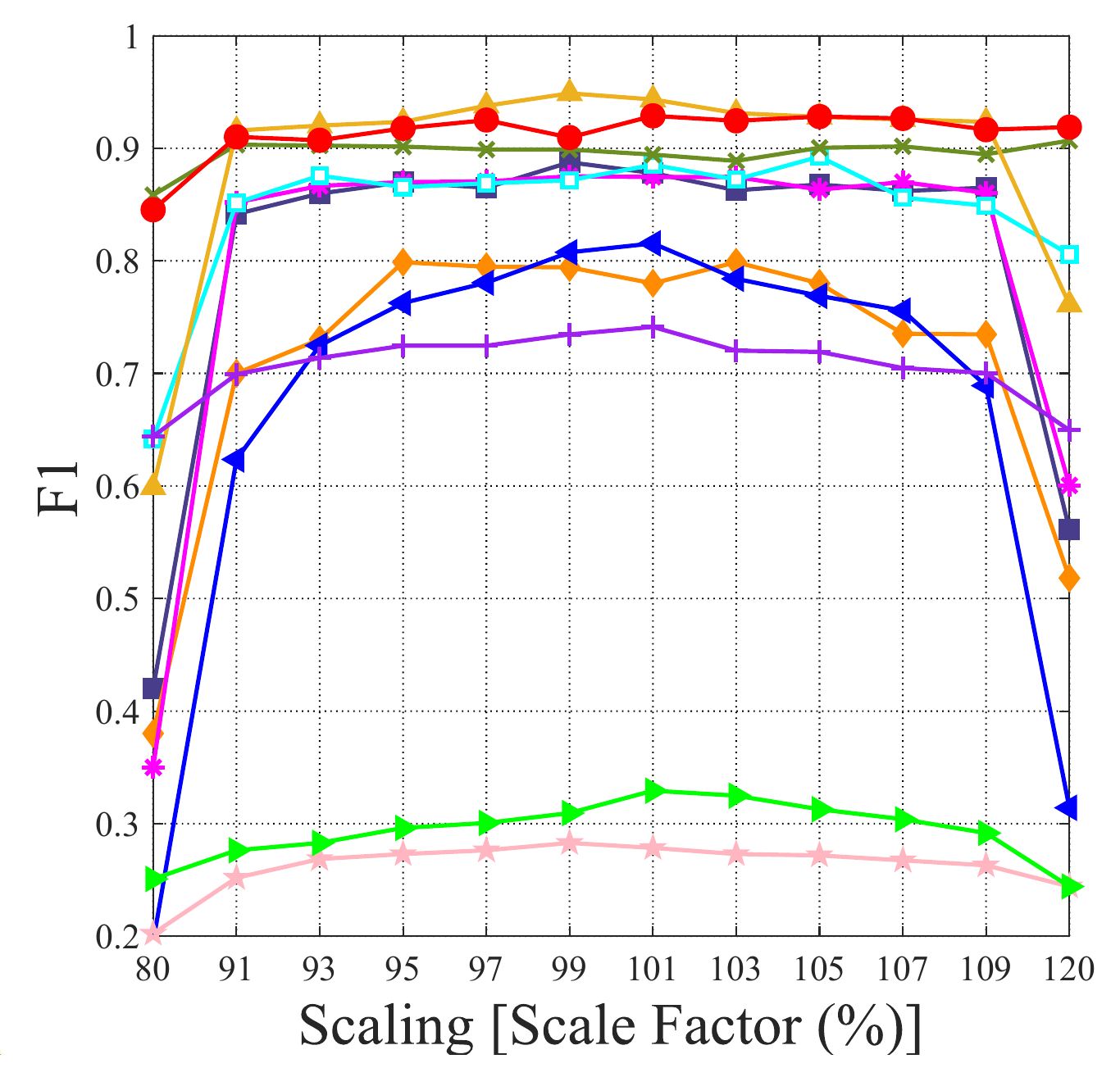}}
	\\
	\subfigure{\includegraphics[width=3in]{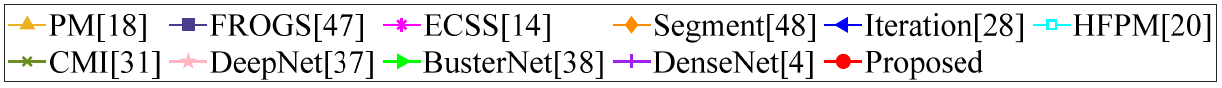}}
	\centering
	\caption{Precision, Recall, and F1 curves for different methods on FAU with various attacks.}
\end{figure*}

\emph{Classical Dense and Sparse Approaches.} Generally, the dense PM, FROGS, and ECSS are more accurate in most light attacks, while the sparse Iteration and HFPM show significant performance advantages for some strong attacks. Here, we observe that the dense methods may tend to fail in the following cases. 1) The curves of FROGS and ECSS drop sharply for rotation of angle $> 20^{\circ}$, while the PM is well designed and does not involve this problem. 2) The curves of all dense methods here, also the PM, drop sharply for scaling of factor $1 \pm > 10\%$. Obviously, such common flaws in geometric invariance can easily form an effective anti-forensic strategy. For sparse methods, although not as accurate as PM in most cases, they can provide more stable results at relatively high level, with no commonality failure being observed.

\emph{Deep Learning Approach.} Similar to the benchmark experiments, the scores of the deep learning methods, especially DeepNet and BusterNet, have not yet reached the level of hand-crafted methods under most attacks. We observe that DenseNet provide quite flat F1 curves at the level of $\sim 70\%$, even in the scaling scenarios where PM drops sharply, implying its critical potential for further research. 

\emph{Our Approach.} In general, our approach inherits the good robustness of sparse methods like Iteration and HFPM (especially for scaling and rotation) while providing a level of performance comparable to dense methods like PM, FROGS, and ECSS. The above facts illustrate that our algorithm can offer better overall robustness accuracy. A noteworthy accuracy decrease is observed at the 60 and 180 degree rotations. The possible reason for this lies in the imperfection of the adaptive clustering. In the future, it is expected to solve this problem by designing more discriminative representation of matched pairs for clustering. In the scaling scenarios, the proposed algorithm exhibits a significant advantage over the existing state-of-the-art works. This should be mainly attributed to the proposed feature and post-processing strategies, both of which have satisfactory robustness. It is worth noting that scaling causes changes in the density of keypoints and rotation causes changes in the order of keypoints. The proposed phrase-level feature descriptions (i.e., geometric phrase pooling and spatial weighting) are adaptive to both changes, thus maintaining similar curves w.r.t. word-level CMI under geometric transformations. Note that the work in this paper does not show a clear advantage w.r.t. CMI and PM (except for scaling) on small-scale FAU with stand-alone and restricted attacks. As illustrated later, the performance of both CMI and PM degrades significantly on large-scale datasets with composite and/or strong attacks.

\subsection{Performance of Key Components}

In this round of experiments, we provide the performance statistics of the key components in our algorithm, verifying their effectiveness directly. Also, this part can be considered as a component-level comparison with previous CMI, where CMI is equipped with 2NN matching, word-level description, and non-fusion post-processing.

\begin{figure}[!t]
	\centering
	\includegraphics[width=1.8in]{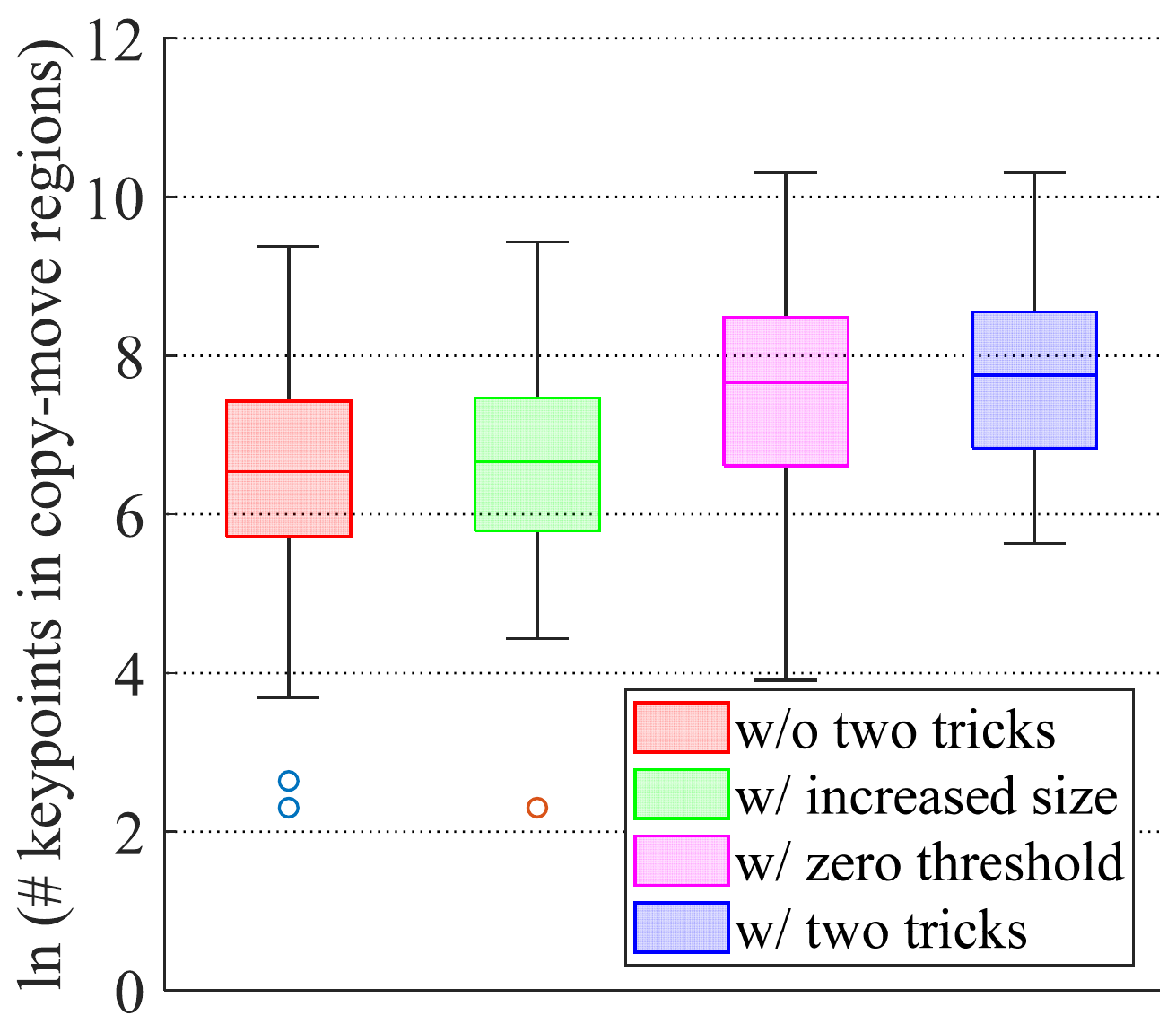}
	\caption{The ablation study for different keypoint detection strategies.}
\end{figure}

\emph{Keypoint Detection.} We first conduct an ablation study for keypoint detection discussed in Section \uppercase\expandafter{\romannumeral3}-B, by counting the number of keypoints in copy-move regions. With different strategies, the distribution for such numbers on FAU is presented in Fig. 11. Note that the natural logarithmic coordinate is used for the $y$-axis. As can be seen, the naive detection without two tricks may cause the following failure: too few keypoints in copy-move regions to support forensic analysis. The introduction of a separate trick, i.e., zero threshold or increased size, somewhat alleviates above problem, but there are still some failure cases. While when both tricks are used, there are $\sim 600$ keypoints in copy-move regions for the worst case, which is generally sufficient for forensic analysis.

\begin{table*}
	\caption{The Ablation Study for Different NN Testing Strategies by Statistics of Efficiency and Matches on FAU Dataset}
	\centering
	\begin{tabular}{ccccc}
		\toprule
		NN Testing & \tabincell{c}{2NN \cite{ref31}} & \tabincell{c}{G2NN \cite{ref32}} & \tabincell{c}{RG2NN \cite{ref18}} & I2NN \\
		\midrule
		CPU-time in second & 76.8 & 148.6 & 279.4 & 85.7 \\ 
		Avg. \#correct matches & 2691.8 & 5318.2 & 5590.6 & 5340 \\
		Min. \#correct matches & 195 & 1813 & 1848 & 1793 \\ 
		Avg. (\#correct matches / \#all matches) \% & 87.8 & 81.9 & 78.1 & 87.9  \\
		Min. (\#correct matches / \#all matches) \% & 65.4 & 57.2 & 52.3 & 67.6  \\
		\bottomrule
	\end{tabular}
\end{table*}

\emph{NN Testing.} In Table \uppercase\expandafter{\romannumeral5}, we conduct an ablation study for different NN testing strategies discussed in Section \uppercase\expandafter{\romannumeral3}-C. Here, the statistics are formed on all the images with multiple copy-move forgeries in FAU. We use the direct NN searching to better reflect the accuracy and efficiency of NN testing strategy itself. Note that the number of correct matches (in both average and minimum) derived by the 2NN strategy is significantly less than the others, for its failure to deal with the multi-feature matching problem. In contrast, the G2NN, RG2NN, and I2NN yielded similar numbers of correct matches, implying that they can fulfill the multi-feature matching to varying degrees. However, the G2NN and especially the RG2NN strategies produce a considerable number of false matches, resulting in a small percentage of correct matches in the set of matched pairs. Such fact obviously increases the difficulty of the post-processing, potentially leading to false positives in the localization result. On the contrary, the proposed I2NN exhibits the highest percentage of correct matches w.r.t. all matches, also the number of correct matches remains at the same level as G2NN and RG2NN, in both average and minimum. Consequently, our I2NN is more capable in both overcoming false matches and dealing with multi-feature matching.

\begin{table}
	\caption{The Ablation Study for Different Feature Description/Matching Strategies by Statistics of Efficiency and Matches on FAU Dataset}
	\centering
	\begin{tabular}{ccc}
		\toprule
		Feature description/matching phases & Word-level & Phrase-level \\
		\midrule
		CPU-time in second & 59.3 & 18.6  \\
		Avg. (\#correct matches / \#all matches) \% & 50.1 & 86.3  \\ 
		Min. (\#correct matches / \#all matches) \% & 4.4 & 36.0  \\
		\bottomrule
	\end{tabular}
\end{table}

\emph{Feature Description/Matching Phases.} In Table \uppercase\expandafter{\romannumeral6}, we conduct an ablation study for different feature description/matching levels (i.e., word and phrase levels) discussed in Sections \uppercase\expandafter{\romannumeral3}-C and \uppercase\expandafter{\romannumeral3}-D. Here, the statistics are formed on all the images in FAU. It can be seen that the proposed phrase-level feature description/matching strategy can significantly exclude the false matches in the word-level, and the time cost of the phrase-level strategy is only one-third of the word-level strategy. Especially for the worst case, the percentage of correct matches w.r.t. all matches is much higher in the phrase-level than in the word-level. Certainly, it must be acknowledged that the above process may remove some correct matches of the word-level. However, the deleted matches are mainly false matches, and the loss of a small number of correct matches is acceptable.

\begin{table}
	\caption{The Ablation Study for Different Post-processing Strategies by Statistics of Localization Accuracy on FAU Dataset}
	\centering
	\begin{tabular}{cccc}
		\toprule
		\tabincell{c}{Post-processing} &\tabincell{c}{Homography and \\local similarity} &\tabincell{c}{Matched \\keypoint regions} & \tabincell{c}{Fusion} \\
		\midrule
		Avg. Precision \% & 67.3 & 47.9 & 93.5  \\
		Avg. Recall \% & 62.3 & 96.9 & 92.8  \\
		Avg. F1 \% & 58.9 & 60.4 & 91.3  \\
		\bottomrule
	\end{tabular}
\end{table}

\emph{Post-processing.} In Table \uppercase\expandafter{\romannumeral7}, we conduct an ablation study for different post-processing strategies discussed in Section \uppercase\expandafter{\romannumeral3}-E. It can be seen that the localization based on matched keypoint regions exhibits low precision. Since such matched regions are only a rough approximation of forgeries. As for the localization strategy based on homography and local similarity, the correlation coefficients may be quite high in similar-but-genuine backgrounds, while being sensitive to some subtle differences in copy-move regions, and thus both the precision and recall are limited. In contrast, the proposed fusion strategy makes full use of the relevant information from homography, local similarity, and keypoint region, thus providing more satisfactory localization results.

\subsection{Performance for Challenging Scenarios}

\begin{table*}[t]
	\caption{Precision, Recall, and F1 scores (\%) for Comprehensive Competitors on CASIA Dataset.}
	\centering
	\begin{threeparttable}
		\begin{tabular}{cccccccccccc}
			\toprule
			Metrics & \tabincell{c}{Zernike \\ \cite{ref11}} & \tabincell{c}{PM \\ \cite{ref17}} & \tabincell{c}{ECSS \\ \cite{ref13}}  & \tabincell{c}{MSA \\ \cite{ref34}} & \tabincell{c}{Segment \\ \cite{ref42}} & \tabincell{c}{HPFM \\ \cite{ref19}} & \tabincell{c}{CMI \\ \cite{ref30}} & \tabincell{c}{DeepNet\\ \cite{ref45}} & \tabincell{c}{BusterNet \\ \cite{ref46}} & \tabincell{c}{DenseNet \\ \cite{ref4}}  & Proposed \\
			\midrule
			Precision & 22.7  & 47.3  & 42.0  & 56.7  & 43.0  & 57.8  & 68.0  & 24.0  & 55.7  & 70.9  & 82.7   \\
			Recall & 13.4  & 49.5  & 45.9  & 54.3  & 36.7  & 65.9  & 88.3  & 13.8  & 43.8  & 58.9  & 81.7   \\
			F1 & 16.4  & 48.4  & 43.9  & 55.5  & 39.6  & 61.6  & 73.3  & 17.5  & 45.6  & 64.3  & 79.6   \\
			\bottomrule
		\end{tabular}
	\end{threeparttable}
\end{table*}

\begin{table*}[t]
	\caption{Precision, Recall, and F1 scores (\%) for Comprehensive Competitors on CoMoFoD Dataset.}
	\centering
	\begin{threeparttable}
		\begin{tabular}{ccccccccccc}
			\toprule
			Metrics & \tabincell{c}{Zernike \\ \cite{ref11}} & \tabincell{c}{PM \\ \cite{ref17}} & \tabincell{c}{MSA \\ \cite{ref34}} & \tabincell{c}{Segment \\ \cite{ref42}}  & \tabincell{c}{HFPM \\ \cite{ref19}} & \tabincell{c}{CMI \\ \cite{ref30}} & \tabincell{c}{DeepNet\\ \cite{ref45}} & \tabincell{c}{BusterNet \\ \cite{ref46}} & \tabincell{c}{DenseNet \\ \cite{ref4}} & Proposed \\
			\midrule
			Precision & 33.6 & 36.9 & 6.7 & 8.9 & 42.8 &  47.1 & 31.6 & 40.4 & 46.1 & 57.2 \\
			Recall & 32.5 & 38 & 6.6 & 8.8 & 42.4 & 63.1 & 29.6 & 33.3 & 42.2 & 53.6 \\ 
			F1 & 33.0 & 37.4 & 6.7 & 8.9 & 42.6 & 48.3 & 30.6 & 36.5 & 44.1 & 50.7\\ 
			\bottomrule
		\end{tabular}
	\end{threeparttable}
\end{table*}

\begin{figure*}[!t]
	\centering
	\includegraphics[width=14cm]{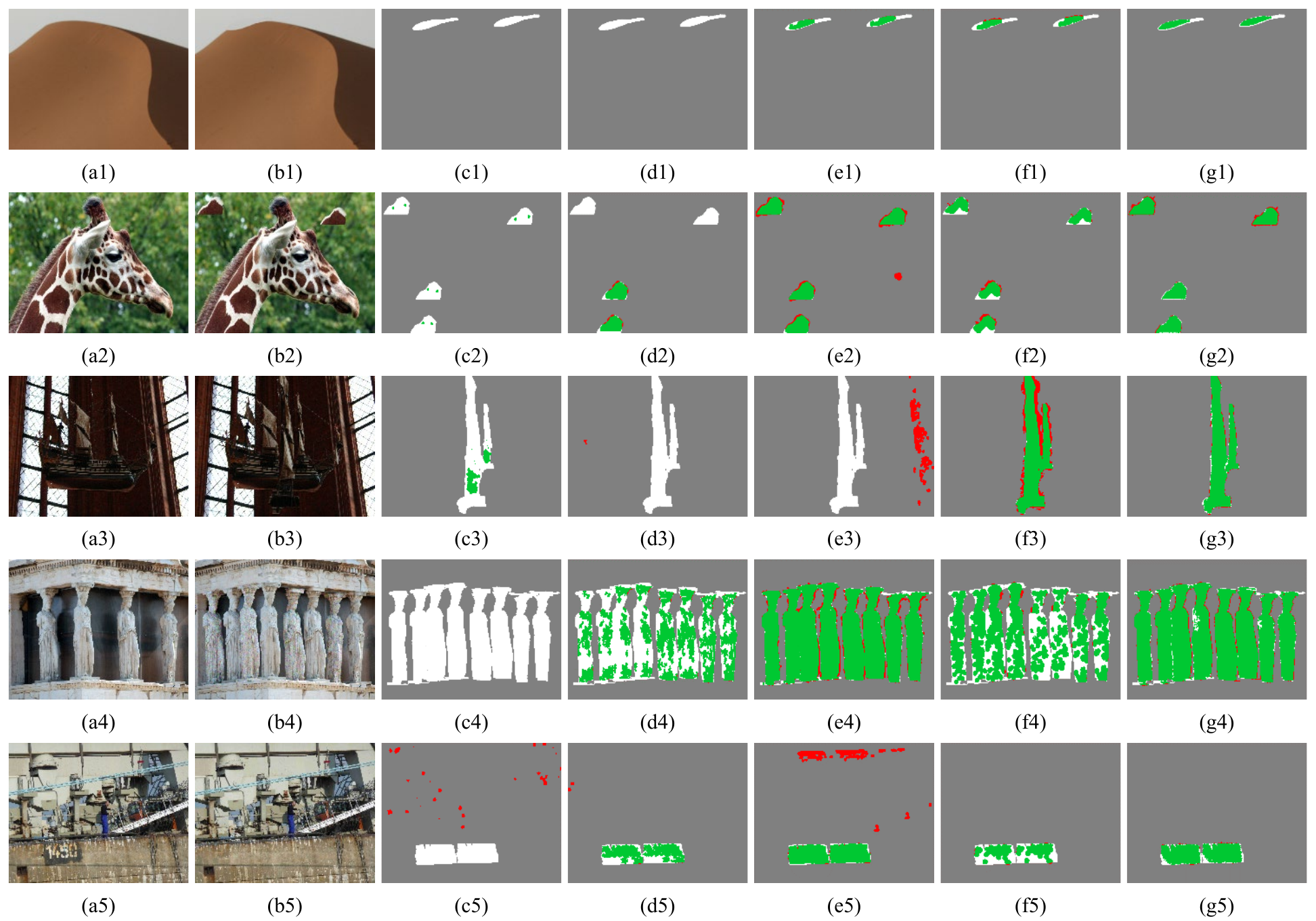}
	\caption{Several challenging samples of the copy-move forgery detection with small/smooth regions, multiple forgeries, large scaling, severe noise, and JPEG compression (from top to bottom). Here, (a1)–(a5) are the original images, (b1)–(b5) are tampered images, (c1)–(c5) are the results for Segment, (d1)–(d5) are the results for PM, (e1)–(e5) are the results for Iteration, (f1)–(f5) are the results for HFPM, and (g1)–(g5) are the results for our work. For (c)–(g), the true positives, false positives, and false negatives are marked in green, red, and white, respectively. The results exhibit that the proposed method provides satisfactory performance for the above challenging samples.}
\end{figure*}

In this round of experiments, we provide in-depth studies under challenging scenarios for comparing our method with state-of-the-art methods, especially the main competitors PM and the previous CMI.

\emph{Challenging Datasets.} In Table \uppercase\expandafter{\romannumeral8}, we provide a score comparison on the challenging dataset CASIA. In addition to the large scale, CASIA is very challenging due to the composite attack, which is closer to the real-world forgery actions than the stand-alone attack of Section \uppercase\expandafter{\romannumeral4}-C. Specifically, for the forgery regions, the pre-processing involves rotation, scaling, and more complex distortions, along with blurring as post-processing. Note that multiple forgeries and similar-but-genuine backgrounds are also involved. Here, the scores of competing methods are cited directly from the reference \cite{ref4}. From the table, one can note a significant decrease in scores for almost all methods w.r.t. the previous experiments. In general, the largest decrease occurs in dense approach, e.g., the PM, due to the lack of geometric invariance (especially for scaling). As expected, sparse approach, e.g., HFPM and CMI, exhibits better scores on challenging CASIA for their good geometric invariance. Meanwhile, in such more complex scenario, deep learning approach especially DenseNet provides promising scores, thanks to the strong adaptiveness for complex data variations. The proposed method achieves the best F1 and Precision scores as well as the second best Recall score, being much better than the state-of-the-art PM, HFPM and DenseNet within three approaches, also outperforming the previous CMI.

In Table \uppercase\expandafter{\romannumeral9}, a similar evaluation is performed on the challenging CoMoFoD dataset. In addition to covering the attacks in CASIA, the CoMoFoD also involves a wider range of post-processing, such as compression, noise, and brightness/color/contrast adjustments, and with nearly four times the size of CASIA. In this more challenging scenario, we observe a similar phenomenon: the scores of the dense approach decrease sharply, while the sparse and deep learning approaches show a well consistency, w.r.t. previous scores in Sections \uppercase\expandafter{\romannumeral4}-B and C. In general, our proposed method still achieves a performance advantage over state-of-the-art PM, HFPM, DenseNet, and  CMI. We also note that the CMI consistently yields better Recall values. More broadly, there is a specific trade-off between Precision and Recall, fundamentally reflecting the contradiction between discriminability and robustness of representation. Compared to our previous CMI, the proposed method introduces mainly enhancement designs for discriminability, rather than robustness.

Therefore, the above common phenomena confirm the validity of numerous techniques and ideas presented in this paper, implying potential applications in challenging real-world forensic scenarios.

\emph{Challenging Cases.} In Fig. 12, some challenging samples are illustrated, where the competing algorithms Segment \cite{ref42}, PM \cite{ref17}, Iteration \cite{ref27}, and HFPM \cite{ref19} are covered. 

    \begin{itemize}
	\item For (a1)–(g1), the small/smooth copy-move regions are involved. It can be seen that Segment and PM fail completely. For the former, the main reason lies in the fact that not enough keypoints are detected in such smooth regions. As for the latter, the random nature of PM algorithm makes it difficult to handle small regions. The Iteration and HFPM successfully detect the copy-move regions, but have a high false-negative rate, as the imperfection of their post-processing algorithms. The proposed method, with well-designed keypoint detection (Section \uppercase\expandafter{\romannumeral3}-B) and post-processing (Section \uppercase\expandafter{\romannumeral3}-E) algorithms, thus exhibits better accuracy for this challenging case. 
	
	\item For (a2)–(g2), the multiple copy-move forgeries are involved. It can be seen that Segment and PM are also fail here, possibly for the similar reasons as before. The Iteration and HFPM are able to detect all the four regions, but both exhibit high false-positive rate and high false-negative rate, respectively. For the Iteration, its absolute distance based NN testing (Section \uppercase\expandafter{\romannumeral3}-C) allows a well multi-feature matching, also leading to low discriminability for similar-but-genuine regions. For the HFPM, its 2NN strategy (Section \uppercase\expandafter{\romannumeral3}-C) should be primarily responsible for the false negatives, as we shown in Section \uppercase\expandafter{\romannumeral4}-D. In contrast, the proposed method works better in this challenging case, further confirming the effectiveness of our I2NN (Section \uppercase\expandafter{\romannumeral3}-C). 
	
	\item For (a3)–(g3), the large scaling is involved. It can be seen that Segment, PM, and Iteration all fail, due to the fact that the descriptors they construct are not scaling invariant. The HFPM and our method successfully detect such copy-scale-move regions. The HFPM exhibits a high false-positive rate, arising from the defects of its matched-region based localization algorithm. As for the proposed method, both word-level and phrase-level features are designed towards rotation and scaling invariance, and hence it able to handle this challenging case. In addition, our fusion-based algorithm, using the information from homography, local similarity, and keypoint region, allows for higher accuracy. 
	
	\item For (a4)–(g4) and (a5)–(g5), severe noise and JPEG compression are considered, respectively. Under such signal corruptions, the localization maps of these comparison methods exhibit varying degrees of inaccuracy, i.e., they are with notable false positives and/or false negatives. In general, the poor robustness/discriminability of their low-level features fundamentally contribute to the above phenomenon. Instead, we focus on building more expressive features through the BoVW model (Section \uppercase\expandafter{\romannumeral3}-D), with clear advantages in overcoming such challenging case as shown in Fig. 12.
	
\end{itemize}

\section{Conclusion}

Although copy-move is a basic and common operation in image forgery, detecting such manipulations can be very difficult, especially for the scenarios with high self-similarity or strong corruption. The fundamental difficulty lies in the following semantic gap problem: features should be robust to geometric/signal distortions in copy-move regions, while retaining discriminability for similar-but-genuine regions. The low-level visual representation employed in existing copy-move forgery detection algorithms is inherently constrained by the above semantic gap problem.

The algorithm designed here is a very early step towards bridging the semantic gap in copy-move forensics. In this regard, we first introduce the bag-of-visual-words model into this field, as a new perspective for the representation in copy-move forgery detection. With this perspective, we then propose a word-to-phrase feature description and matching pipeline, covering the spatial structure and visual saliency information of images. The core lies in a spatial pooling and weighting of local moment invariants for robust and discriminative representation, which is expected to alleviate the semantic gap. In addition to the above core contributions, we also give some useful modules that can complement the existing works. Specifically, our I2NN is able to deal with multiple copy-move forgeries, with much fewer mismatches and complexity; our fusion-based localization strategy using the information from homography, local similarity, and keypoint region, with ability to reduce both false positives and false negatives.

We provide statistical comparisons with both state-of-the-art dense-field and sparse-field forensics methods, by both benchmark experiments and robustness experiments. In general, the extensive experimental results confirm our claims, exhibiting quite satisfactory accuracy and robustness w.r.t. existing works.

\bibliographystyle{IEEEtran}

\bibliography{ref}
\def\bibfont{\small}





 


\begin{IEEEbiography}[{\includegraphics[width=1in,height=1.25in,clip,keepaspectratio]{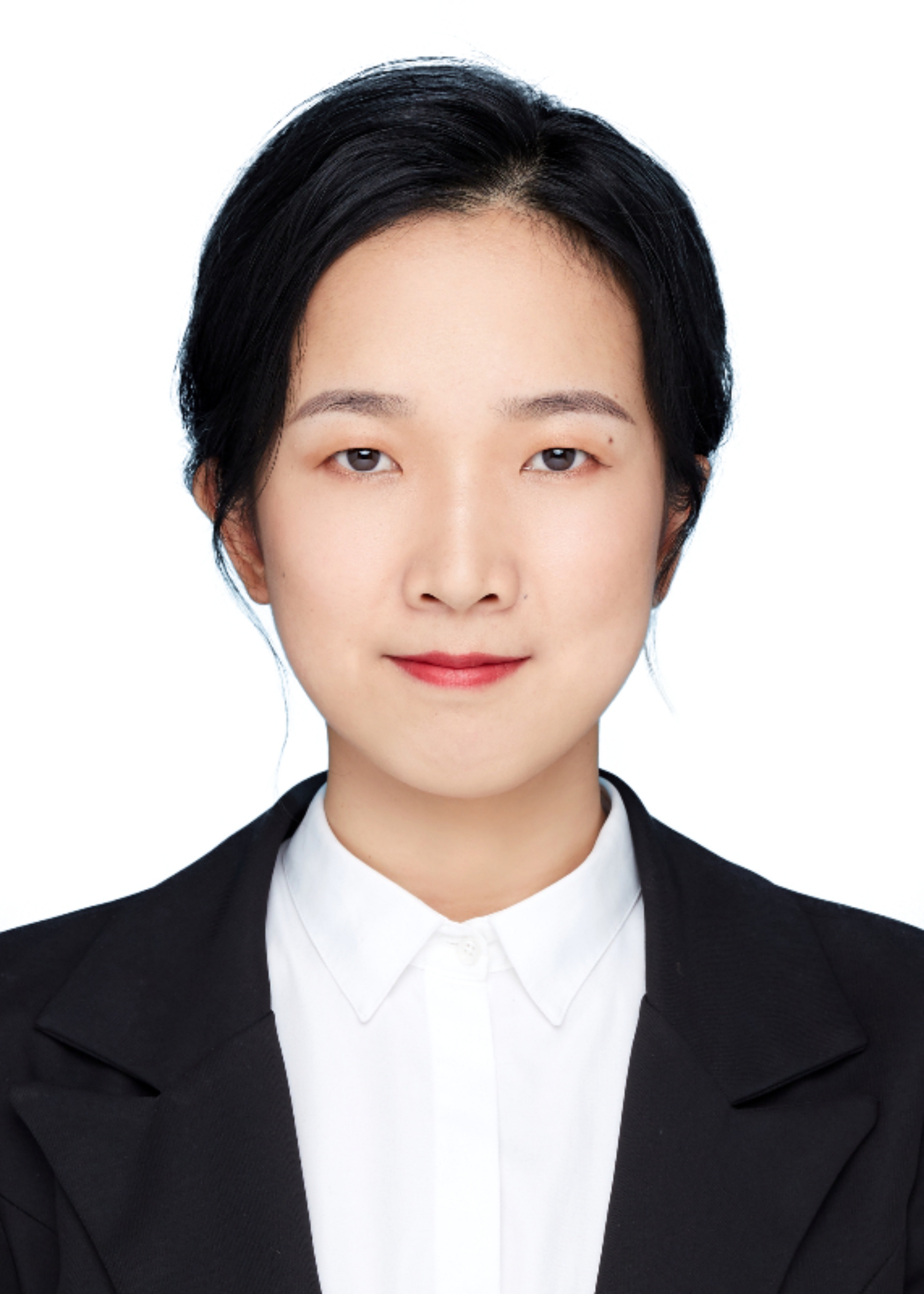}}]{Chao Wang}
	received the B.S. and M.S. degrees from Liaoning Normal University, Dalian, China. She is currently pursuing the Ph.D. degree in computer science at Nanjing University of Aeronautics and Astronautics, Nanjing, China. Her research interests include trustworthy artificial intelligence, adversarial learning, and media forensics.
\end{IEEEbiography}

\begin{IEEEbiography}[{\includegraphics[width=1in,height=1.25in,clip,keepaspectratio]{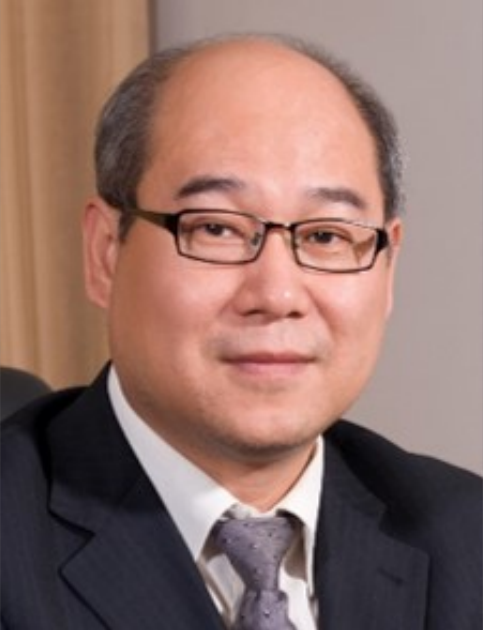}}]{Zhiqiu Huang}
	is a professor in the College of Computer Science and Technology, Nanjing University of Aeronautics and Astronautics, China. He received the Ph.D. degree in computer science from the Nanjing University of Aeronautics and Astronautics in 1999. He has authored over 80 papers in referred journals and proceedings in the areas of software engineering and knowledge engineering. His research interests include software engineering, formal methods, and knowledge engineering.
\end{IEEEbiography}

\begin{IEEEbiography}[{\includegraphics[width=1in,height=1.25in,clip,keepaspectratio]{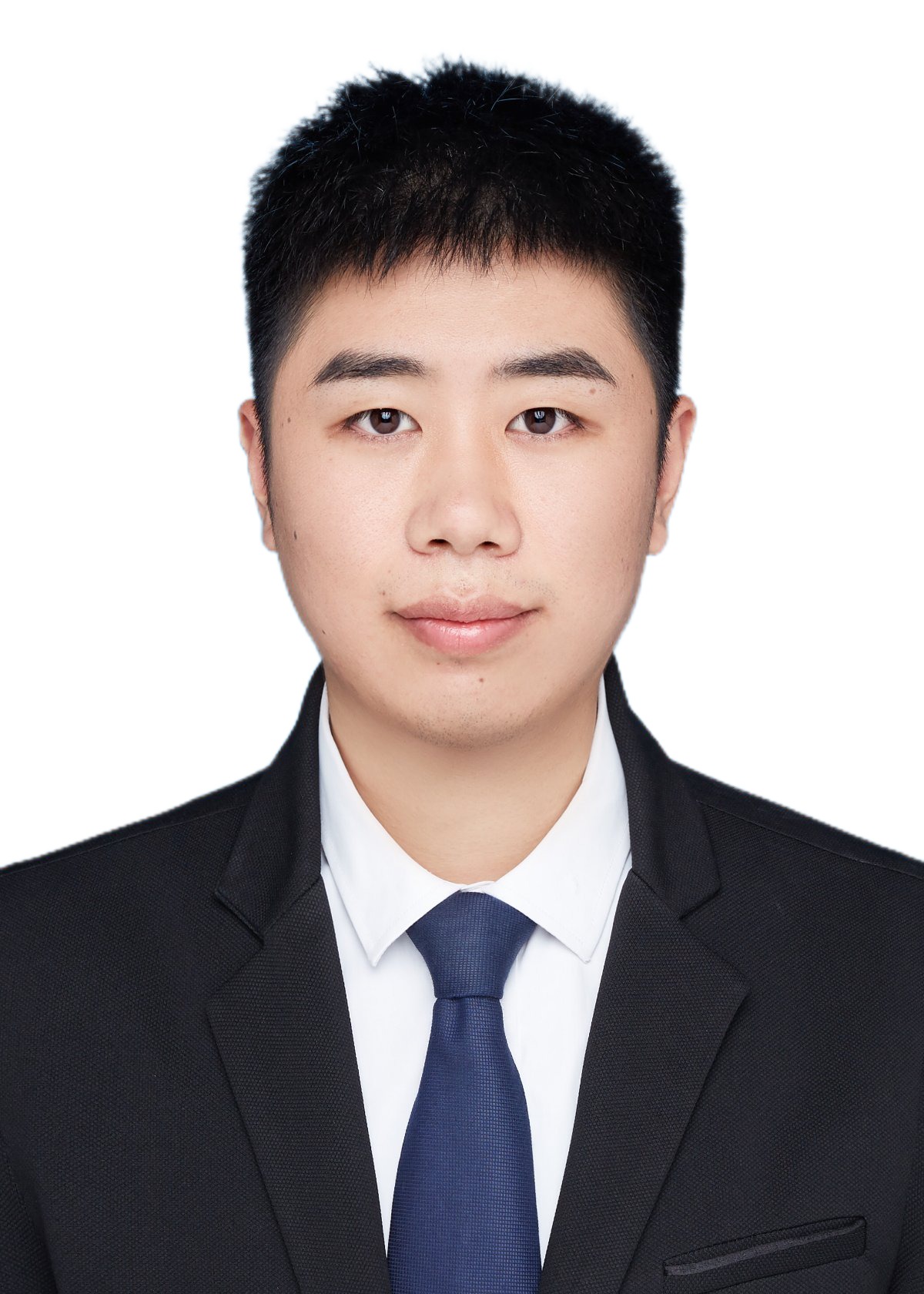}}]{Shuren Qi}
	received the B.A. and M.E. degrees from Liaoning Normal University, Dalian, China, in 2017 and 2020 respectively. He is currently pursuing the Ph.D. degree in computer science at Nanjing University of Aeronautics and Astronautics, Nanjing, China. He has published academic papers in top-tier venues including \emph{ACM Computing Surveys} and \emph{IEEE Transactions on Pattern Analysis and Machine Intelligence}. His research interests include invariant feature extraction and visual signal representation with applications in robust pattern recognition and media forensics/security.
\end{IEEEbiography}

\begin{IEEEbiography}[{\includegraphics[width=1in,height=1.25in,clip,keepaspectratio]{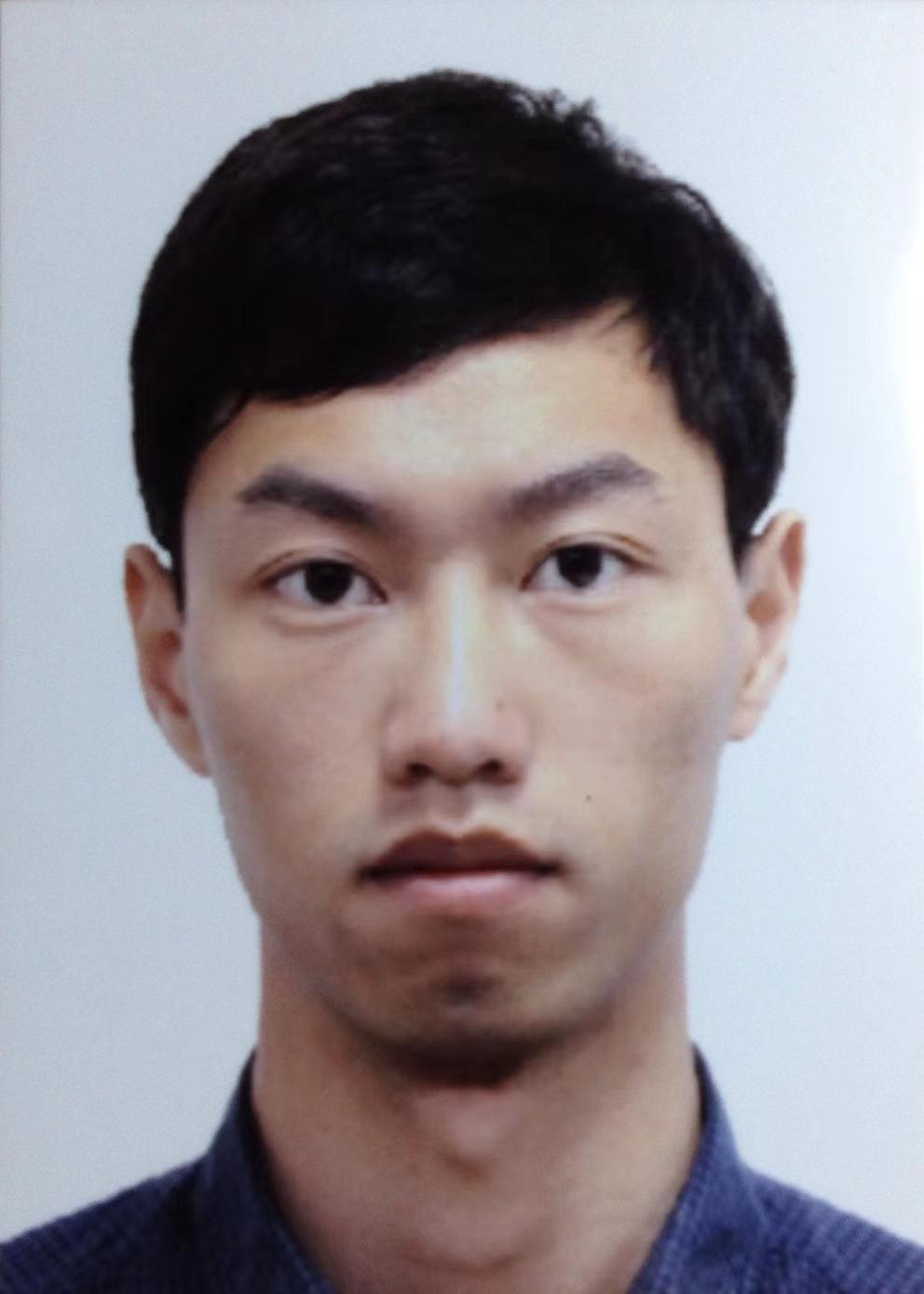}}]{Yaoshen Yu}
	is a Ph.D. candidate in software engineering at the College of Computer Science and Technology, Nanjing University of Aeronautics and Astronautics, China. His research interests include software engineering, code similarity detection, and code recommendation.
\end{IEEEbiography}

\begin{IEEEbiography}[{\includegraphics[width=1in,height=1.25in,clip,keepaspectratio]{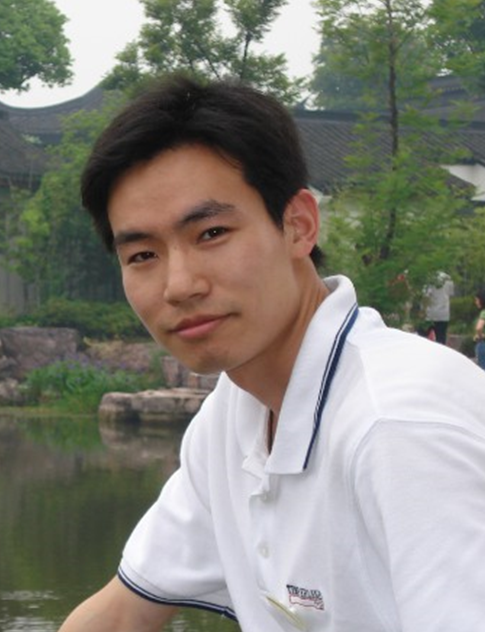}}]{Guohua Shen}
	received the M.S. and Ph.D. degrees in computer science from the Nanjing University of Aeronautics and Astronautics, China. He is currently an Associate Professor with the College of Computer Science and Engineering, Nanjing University of Aeronautics and Astronautics. His research interests include knowledge graph, software quality, and service privacy.
\end{IEEEbiography}

\begin{IEEEbiography}[{\includegraphics[width=1in,height=1.25in,clip,keepaspectratio]{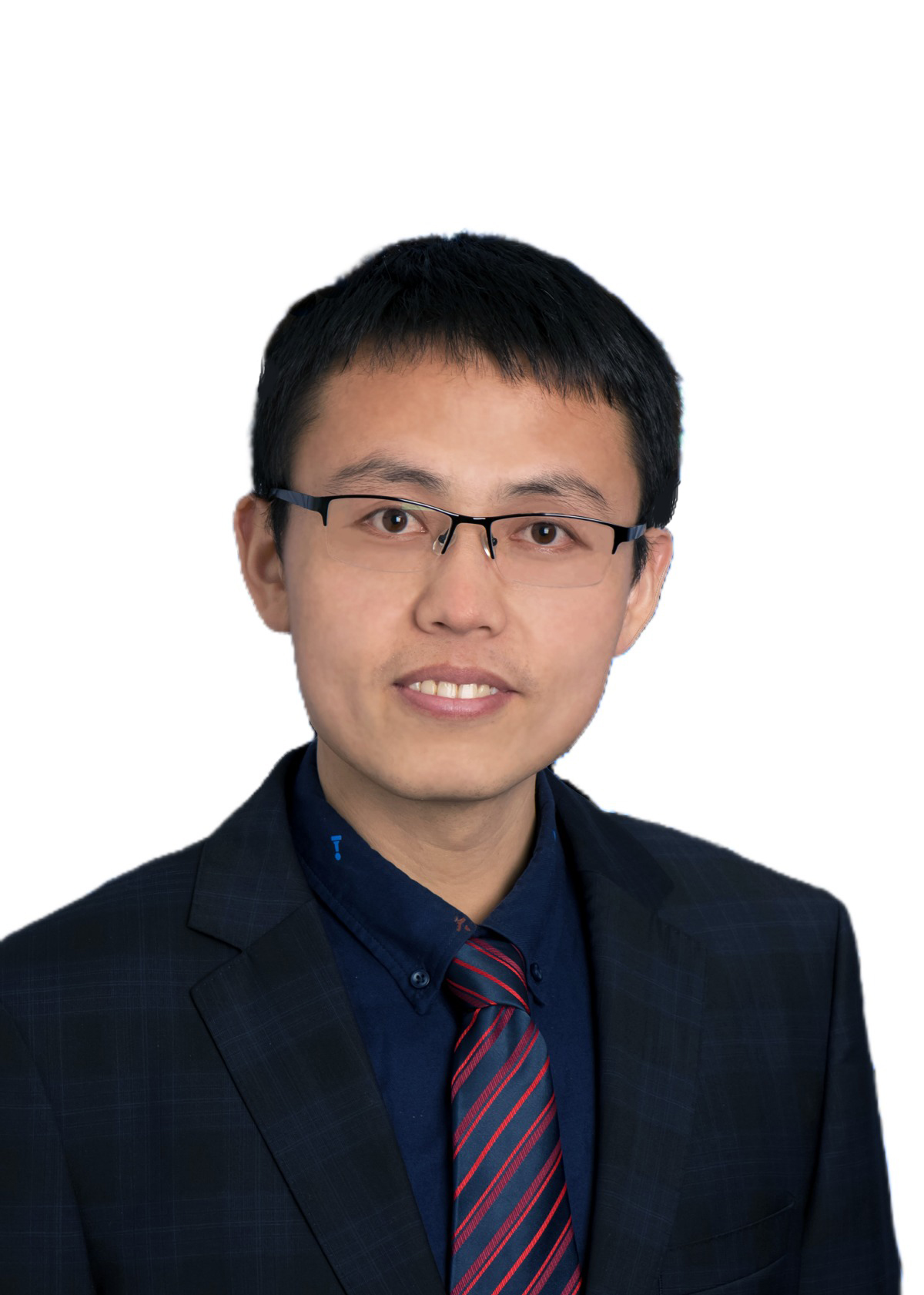}}]{Yushu Zhang}
	(Member, IEEE) received the Ph.D. degree in computer science from Chongqing University, Chongqing, China, in 2014. He held various research positions with the City University of Hong Kong, Southwest University, University of Macau, and Deakin University. He is currently a Professor with the College of Computer Science and Technology, Nanjing University of Aeronautics and Astronautics, Nanjing, China. His research interests include multimedia processing and security, artifi cial intelligence, and blockchain. Dr. Zhang is an Associate Editor of \emph{Signal Processing} and \emph{Information Sciences}.
\end{IEEEbiography}

\vspace{11pt}


\vfill

\end{document}